\documentclass[letterpaper]{article} 
\usepackage[submission]{aaai25}  
\usepackage{times}  
\usepackage{helvet}  
\usepackage{courier}  
\usepackage[hyphens]{url}  
\usepackage{graphicx} 
\usepackage{natbib}  
\usepackage{caption} 
\usepackage{amsmath}
\DeclareMathOperator*{\argmax}{argmax} 
\usepackage{xspace}
\usepackage{graphicx}   
\usepackage{amsthm}
\usepackage{float}
\usepackage{multirow}
\usepackage{lmodern} 
\graphicspath{{figs/}}
\DeclareGraphicsExtensions{.pdf}

\usepackage{algorithm}
\usepackage{algpseudocode}
\usepackage{multicol}
\usepackage{cuted}
\usepackage{lipsum}

\usepackage{caption}
\usepackage{subfigure}
\usepackage{booktabs}

\usepackage[utf8]{inputenc} 
\usepackage[T1]{fontenc}    
\usepackage{url}            
\usepackage{booktabs}       
\usepackage{amsfonts}       
\usepackage{nicefrac}       
\usepackage{microtype}      
\usepackage{xcolor}         

\newcommand{\todo}[1]{\textcolor{orange}{TODO: #1}}

\newcommand{\jiaxiang}[1]{\textcolor{brown}{Jiaxiang: #1}}

\newcommand{\algorithmname}{NeuralMDC\xspace}

\newcommand{\ie}{\textit{i.e.,}\xspace}
\newcommand{\eg}{\textit{e.g.,}\xspace}

\newcommand{\fig}{Fig.~}

\newcommand{\reff}[1]{Fig.~\ref{f:#1}}

\usepackage{listings}
\usepackage{newfloat}
\DeclareCaptionStyle{ruled}{labelfont=normalfont,labelsep=colon,strut=off} 
\floatstyle{ruled}
\newfloat{listing}{tb}{lst}{}
\floatname{listing}{Listing}
\title{Robust Multiple Description Neural Video Codec with Masked Transformer for Dynamic and Noisy Networks}

\author{
    Xinyue Hu,
    Wei Ye,
    Jiaxiang Tang,
    Eman Ramadan, Zhi-Li Zhang
}
\affiliations{
    University of Minnesota Twin Cities, Minneapolis, USA\\

    \{hu000007, ye000094, tang0836\}@umn.edu, \{eman,zhzhang\}@cs.umn.edu
}

\begin{document}

\maketitle
\begin{abstract}
   Multiple Description Coding (MDC) is a promising error-resilient source coding method that is particularly suitable
   for dynamic networks with multiple (yet noisy and unreliable) paths. However, conventional MDC video codecs suffer from cumbersome architectures, poor scalability, limited loss resilience, and lower compression efficiency. As a result, MDC has never been widely adopted. Inspired by the potential of neural video codecs, this paper rethinks MDC design. We propose a novel MDC video codec, \algorithmname, demonstrating how bidirectional transformers trained for masked token prediction can vastly simplify the design of MDC video codec. 
    %
    %
   To compress a video, \algorithmname starts by tokenizing each frame into its latent representation and then splits the latent tokens to create multiple descriptions containing correlated information. Instead of using motion prediction and warping operations, \algorithmname trains a bidirectional masked transformer to model the spatial-temporal dependencies of latent representations and predict the distribution of the current representation based on the past.  The predicted distribution is used to independently entropy code each description and infer any potentially lost tokens. 
    %
    Extensive experiments demonstrate  \algorithmname achieves state-of-the-art loss resilience with minimal sacrifices in compression efficiency, significantly outperforming the best existing residual-coding-based error-resilient neural video codec.

\end{abstract}

\section{Introduction}
Video delivery is integral to many popular Internet applications and has dominated the Internet traffic.  Emerging 5G networks are designed to 
enable new applications such as augmented/virtual/extended reality, tele-operated robots, and remote driving -- all of which rely on streaming pre-recorded or real-time videos. 
5G networks are capable of delivering beyond 1~Gbps of \emph{peak} throughput by leveraging multiple radio signal paths and/or multiple radio channels~\cite{DynSPAN24-Muhamed-Wei, li2023ca++, ye2024dissecting}. 
However, 5G throughput is known to suffer from wild fluctuations due to noisy radio environments and dynamic changes in availability in MIMO (multi-input, multi-output) layers or radio channel conditions~\cite{first5G, narayanan2020lumos5g, narayanan2021variegated, ye2024dissecting}.

Recent studies~\cite{narayanan2021variegated, 5gaware} reveal that video streaming applications underperform in 5G networks. Despite achieving high bitrates in 5G networks, existing streaming systems experience significantly higher stall times due to delays in receiving the necessary packets for successful video decoding, since conventional video codecs are highly sensitive to packet loss\footnote{In this study, packet loss refers to both packets dropped in transit and packets not received by the decoding deadline. Video streaming can experience a loss rate ranging from 0\% to 80\%~\cite{cheng2024grace}.}. 
While forward error correction (FEC)~\cite{wicker1999reed, badr2017fec} and retransmission (RTX)  are implemented to mitigate packet loss, their effectiveness is limited. This limitation arises from difficulties in determining optimal FEC redundancy parameters for dynamic networks in advance, and the significant delays introduced by RTX. Furthermore, bitrate adaptation algorithms~\cite{hu2023corel} have been utilized to adjust codec bitrates in response to throughput fluctuations, yet the significant variability in 5G network throughput poses substantial challenges to their accuracy. 
Moreover, the complexity of distributing packets across heterogeneous network paths adds another layer of complication to video streaming in 5G networks. 
These observations raise the question: \textit{Can we design a ``proactive'' video codec that is inherently robust to noisy networks and that can more effectively utilize multiple network paths, rather than relying solely on the streaming techniques previously mentioned?}

Conventional video codecs such as AVC and  HEVC  as well as the more recent neural codecs~\cite{lu2019dvc, 2022-VCT, 2022-decompose}  compress a video into a single bitstream for network delivery. In contrast, a 
Multiple Description Coding (MDC)~\cite{ kazemi2014review} video codec compresses a video into multiple \textit{independently-decodable} and \textit{mutually-refinable} streams (also called descriptions). Hence, MDC provides a promising alternative paradigm for video delivery over noisy and dynamic networks such as 5G networks, as it makes it possible to dynamically exploit the availability of multiple noisy radio paths or channels. 
For instance, 
each video stream can be transmitted separately over different network paths/channels using various networking mechanisms.  
If one or more path/channel suffer significant impairments or become unavailable, as long as one or more (even partial) descriptions are received, the video can be successfully decoded, albeit with lower fidelity. 

Despite such advantages, designing an efficient MDC video codec is nontrivial.  Existing MDC video codecs~\cite{franchi2005multiple,  le2023multiple} are largely extensions of AVC/HEVC. They suffer from cumbersome architectures, requiring different side decoders for each description and a central decoder for combined descriptions. This results in poor scalability when generating more than 2 MDC streams. 
They also exhibit limited loss resilience due to the de-correlated nature of DCT transforms and extremely complicated encoder-decoder state synchronization. 
To improve loss resilience, existing MDC techniques often oversample or duplicate source information, resulting in lower compression efficiency. Consequently, MDC  codecs have never been widely adopted in practice.

Inspired by the rapid advances in neural video compression, which outperforms AVC and HEVC in rate distortion performance~\cite{lu2019dvc, 2022-VCT, 2022-decompose}, in this paper, we rethink the design of MDC through the lens of neural codecs. 
We find that bidirectional transformers~\cite{chang2022maskgit} trained for masked token prediction simplify and enhance MDC design. Our neural MDC codec compresses videos in three steps (see \reff{framework}).
First, we use a lossy AutoEncoder transform to independently map each frame $x_t$ to a quantized representation $y_t$. 
Second, we split $y_t$ into multiple non-overlapping parts to form multiple descriptions. 
Third, a masked transformer extracts spatial and temporal redundancies to model the distributions of $y_t$  conditioned on previous representations. We use these predicted distributions and entropy coding to compress each description independently into a bitstream. 
At the receiver side, the received descriptions are merged into $\widetilde y_t$ and decoded to reconstruct the frame $\hat{x}_t$. If any part of $y_t$ is lost during transmission or fails to arrive before the decoding deadline,  the predicted distributions infer the missing part by leveraging spatial-temporal dependencies among representations. 


To the best of our knowledge, this paper is the first to utilize neural compression to design MDC video codec, making video compression more robust and adaptive to network dynamics. Our \algorithmname video codec is elegantly simple yet powerful, leveraging a masked transformer to capture spatial-temporal correlations and leverage arbitrary relationships between frames. 
Our approach avoids complex state synchronization or warping operations, achieving state-of-the-art loss resilience performance and outperforming the best existing loss-resilient neural video codec, Grace~\cite{cheng2024grace}, by 2 to 8 times in terms of PSNR and MS-SSIM of reconstructed videos with packet losses. Additionally, our \algorithmname achieves 76.88\% bitrate savings over Grace. 
It is particularly suited for 5G networks where one can intelligently and adaptively leverage  multiple radio paths/channels when available, while combating the challenges posed by highly noisy and dynamic radio environments.

\section{Related Work}
\noindent\textbf{MDC codecs.} 
The earliest works~\cite{fleming1999generalized} on MDC design focused on developing various quantizers to ensure each description contains the full source information at different levels of coarseness. This line of research primarily focused on rate-distortion optimization of MDC through theoretical analysis and was mainly pursued within the information theory community. Later, the design of MDC shifted towards splitting source information into multiple descriptions, each containing a portion of the source data. 
Depending on the type of source information used, descriptions are generated by partitioning either the pixels~\cite{shirani2006content, yapici2008downsampling} in the spatial domain, or frames~\cite{tillo2004low,radulovic2009multiple} in the temporal domain, or transformed data~\cite{wang2001multiple,conci2007real} in the frequency domain. 
In recent years, neural networks have been used to enhance MDC design. 
Techniques such as CNNs~\cite{zhao2018multiple}, AutoEncoders~\cite{zhao2022lmdc}, and Implicit Neural Representations~\cite{le2023inr} have been utilized to create MDC image codecs. 
However, little attention has been given to MDC video codecs, with the only work~\cite{hu2021multiple} which proposed a GNN-based super-resolution method to improve the reconstruction quality of a traditional MDC video codec.
   



\noindent\textbf{Neural video codecs.}
Numerous research papers have emerged on neural video codecs.  The authors in~\cite{lu2019dvc} introduce the first end-to-end deep learning model that jointly optimizes all components of the video codec. This model uses learning-based optical flow for motion estimation and frame reconstruction. Subsequent work focuses on simplifying module complexity and training schemes, as well as improving the accuracy of motion estimation. For example, \cite{2022-decompose} decomposes the motion information to better model it; \cite{2020-scale} proposes the generalized warping operator and scale-space flow; and \cite{2021-FVC} utilizes the feature space video coding network. At the same time, \cite{2021-Deep} proposes a context-based conditional coding framework, aiming to achieve higher compression rates than the aforementioned predictive coding framework. Following it, \cite{2022-VCT} uses Transformer to predict the distribution of future frames. 
Inspired by the success of implicit neural representations, another research direction \cite{chen2021nerv, kwan2024hinerv} represents videos as neural networks with frame indices as inputs, significantly improving decoding speed and video quality. 

\begin{figure*}[t]
\vspace{-2mm}
    \centering
    \vspace{-4mm}
    \includegraphics[width=\textwidth]{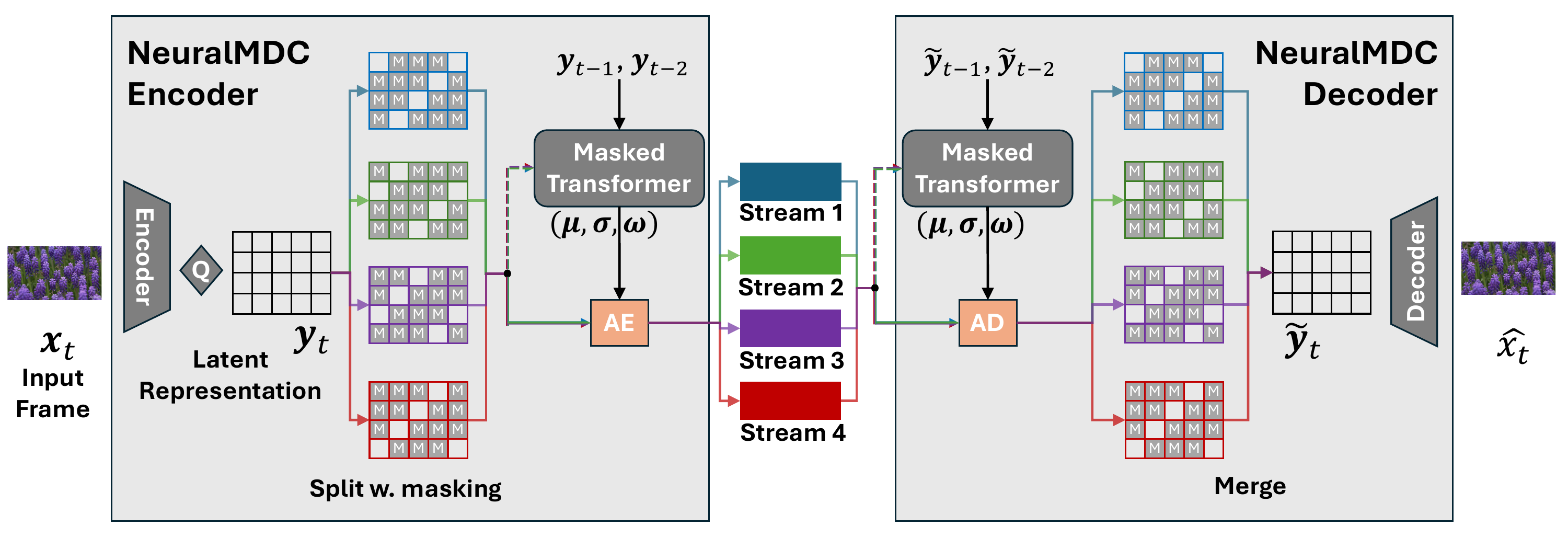}
    \vspace{-7mm}
    \caption{Overview of \algorithmname codec: an example of generating 4 descriptions.} 
   \vspace{-2mm}
    \label{f:framework}
\end{figure*}

\begin{figure*}
    \centering
    \subfigure{\includegraphics[width=0.18\textwidth]{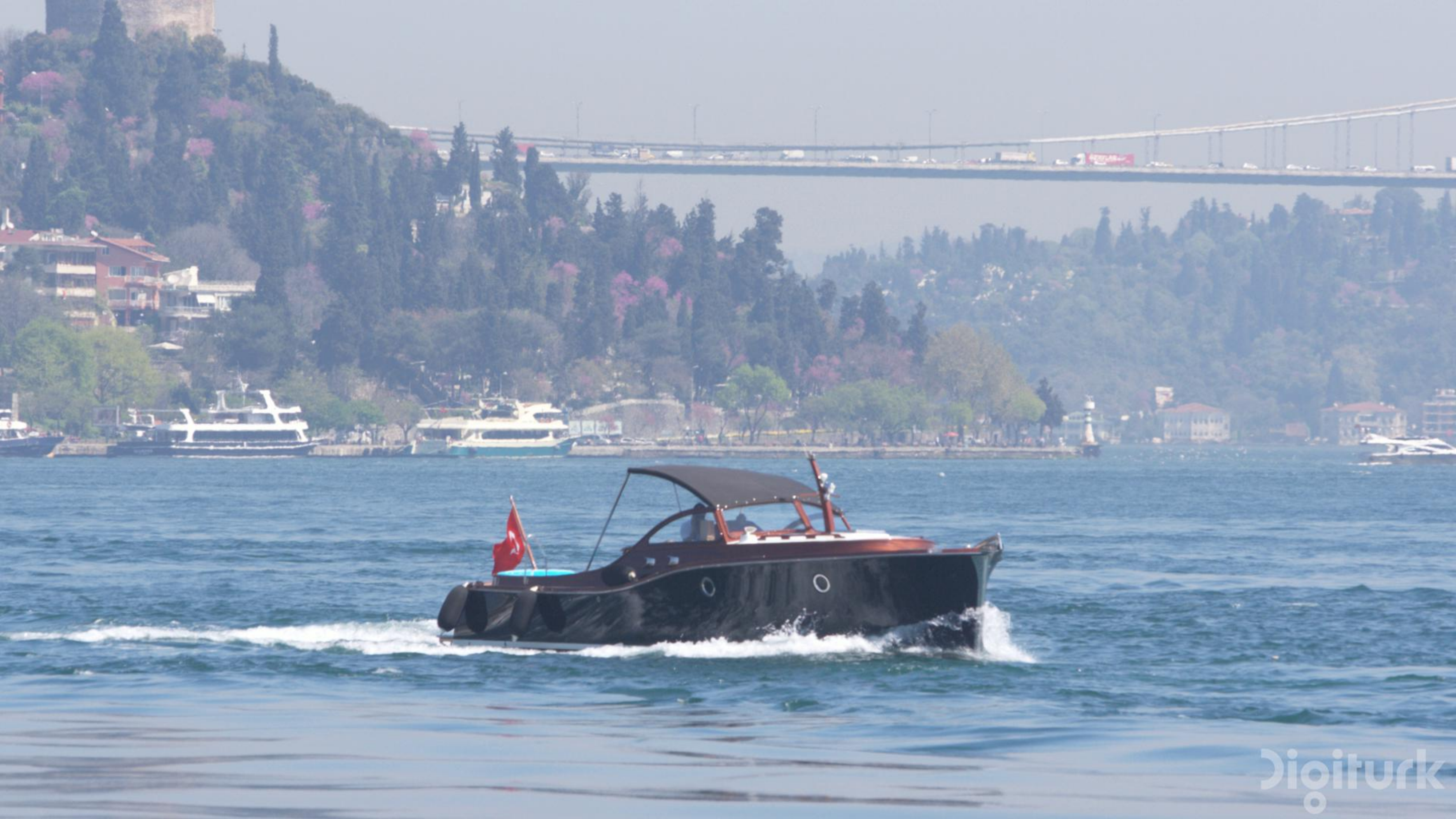}} 
    \subfigure{\includegraphics[width=0.19\textwidth]{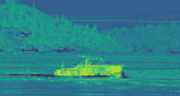}} 
    \subfigure{\includegraphics[width=0.19\textwidth]{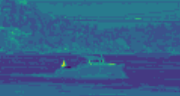}}
    \subfigure{\includegraphics[width=0.19\textwidth]{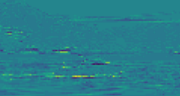}}
    \subfigure{\includegraphics[width=0.19\textwidth]{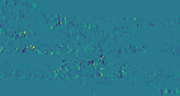}}
    \vspace{-3mm}
   \caption{Sorted channel maps with the top-4 largest energy. Left 1: the original frame. The strongest activation is concentrated in the first channel map (left 2), while the remaining channels become increasingly sparse.}
   \vspace{-4mm}
    \label{fig:channel_maps}
\end{figure*}

\section{Method}\label{s:method}


\subsection{Overview}

In general, the design of an MDC video codec involves three main challenges: 1)~splitting the source information into multiple descriptions that contain correlated (\ie redundant) information, 2)~exploiting redundancy among these descriptions to estimate any potentially lost from those received; and 3)~handling error propagation due to the mismatch of encoder and decoder states.

We use neural compression techniques to design MDC and address the above challenges judiciously. 
A high-level overview of our approach is shown in \fig \ref{f:framework}. 
We generate multiple descriptions in the latent domain using a CNN-based AutoEncoder to tokenize each frame $x_t$ into a quantized latent representation $y_t$. Unlike DCT transforms, which de-correlates the coefficients, AutoEncoder transforms retain spatial-temporal correlations in the latent domain~\cite{he2022elic, li2023mage}. Thus, we split representation $y_t$ into different descriptions containing correlated information. 

To transmit each description with fewer bits as well as to exploit temporal and spatial correlations among descriptions, we use a bidirectional masked transformer to parameterize the distribution of representation $P(y_t | y_{t-1}, y_{t-2})$. 
We then use the predicted distribution and entropy coding (EC) to independently convert each description to a bitstream with $\approx \sum_i -\log_2P(y_{t}^i)$ bits~\cite{minnen2020channel}. If any descriptions are lost, we use the predicted distribution to infer the lost parts. Better distribution prediction results in fewer bits for $y_t$ and more accurate loss inference.
{\em Since each description is entropy encoded independently, each one is independently decodable}. Any combination of received descriptions enhances the decoded latent representation's accuracy and improves the decoded frame's visual quality. 
By avoiding the use of motion vector or warping operations and limiting conditioning to the previous two representations, the impact of temporal error propagation caused by loss is confined to a few local frames.



\subsection{AutoEncoder Transform}
We use an existing CNN-based ELIC AutoEncoder~\cite{he2022elic} to independently convert each input frame from the pixel domain to the latent domain. 
Given an $H\times W$ frame $x_t$, the CNN-based image encoder $E$ maps it to a latent representation of shape $(h, w, c)$, where $h, w$ are $16\times$ smaller than the input resolution and $c$ is set to be $192$ throughout the paper. Following existing works~\cite{lu2019dvc, 2022-VCT}, we quantize the latent representation element-wise using scalar quantization and get the quantized representation $y_t=\lfloor E(x_t) \rfloor $. 
From $y_t$, the decoder $D$ reconstructs the input frame $\hat{x}_t = D(y_t)$. 



%


\begin{figure}[t]
\vspace{-2mm}
\begin{minipage}{.5\textwidth}
    \centering
    \includegraphics[width=\textwidth]{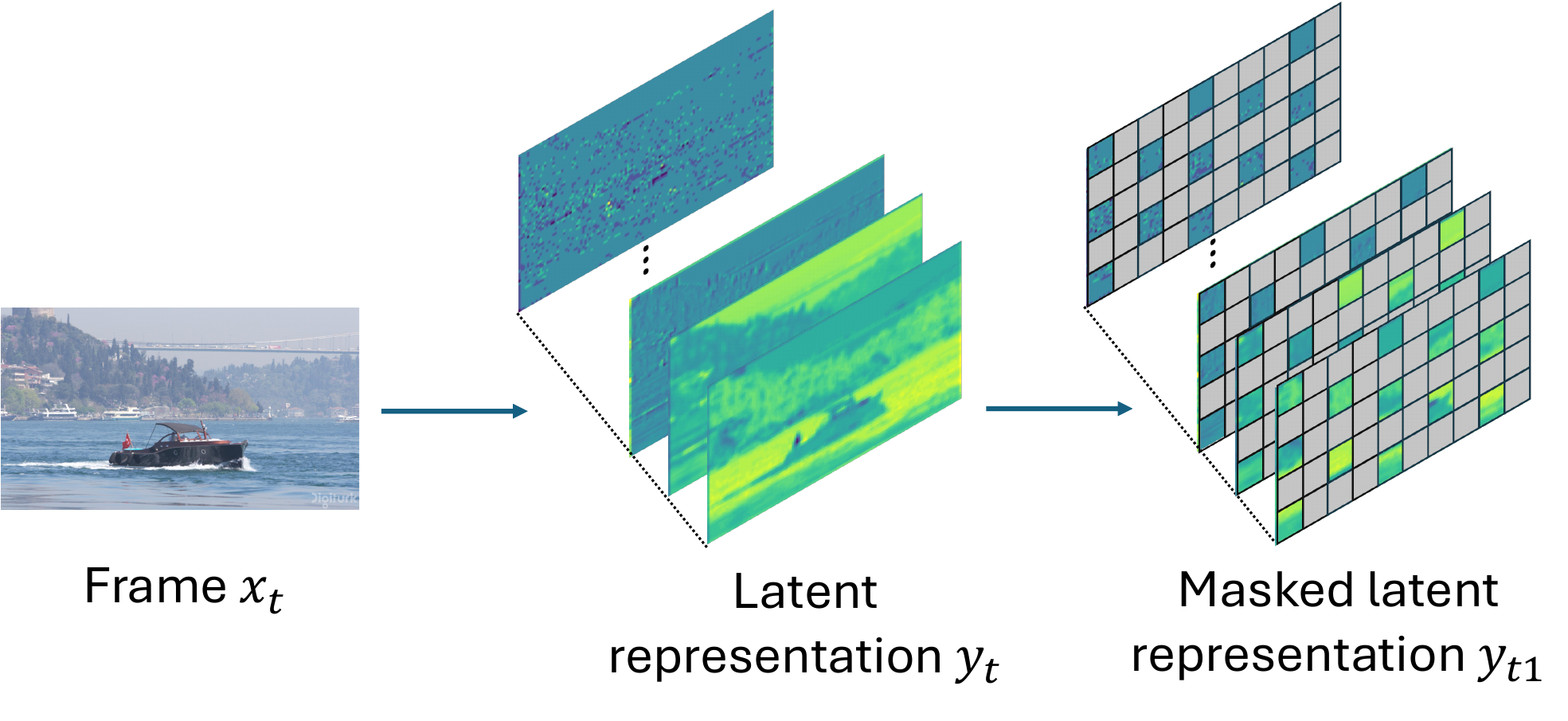}
    \vspace{-6mm}
    \caption{Latent representation separation example: the $1/4$ description.}
    \vspace{-0mm}
    \label{f:seperation_example }
\end{minipage}
\hfill
\begin{minipage}{.45\textwidth}
    \subfigure{\includegraphics[width=.48\linewidth]{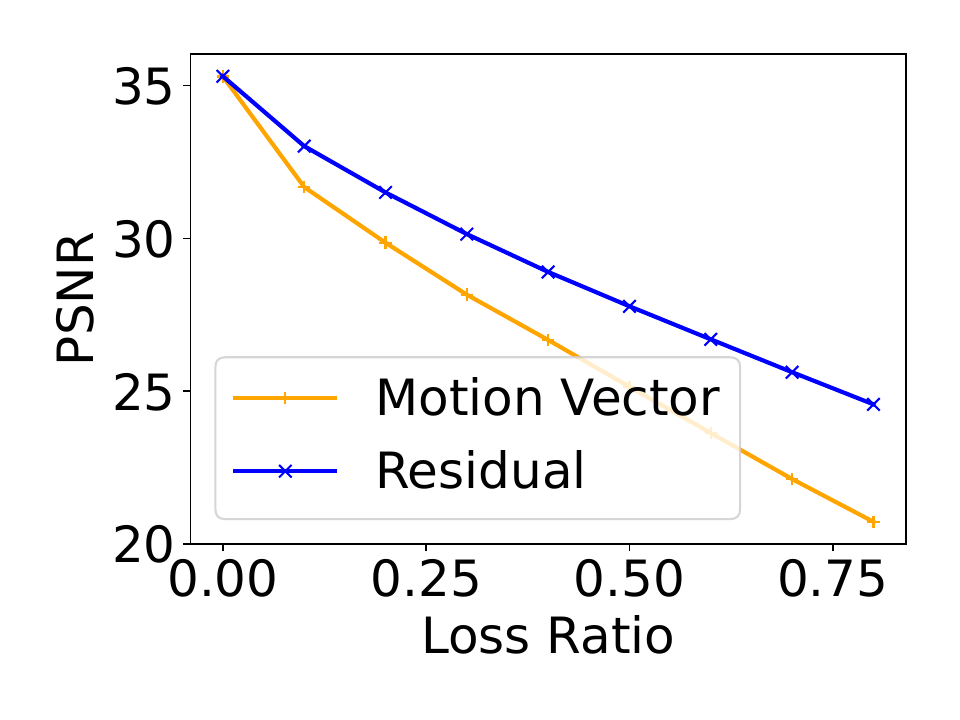}}
    \subfigure{\includegraphics[width=.48\linewidth]{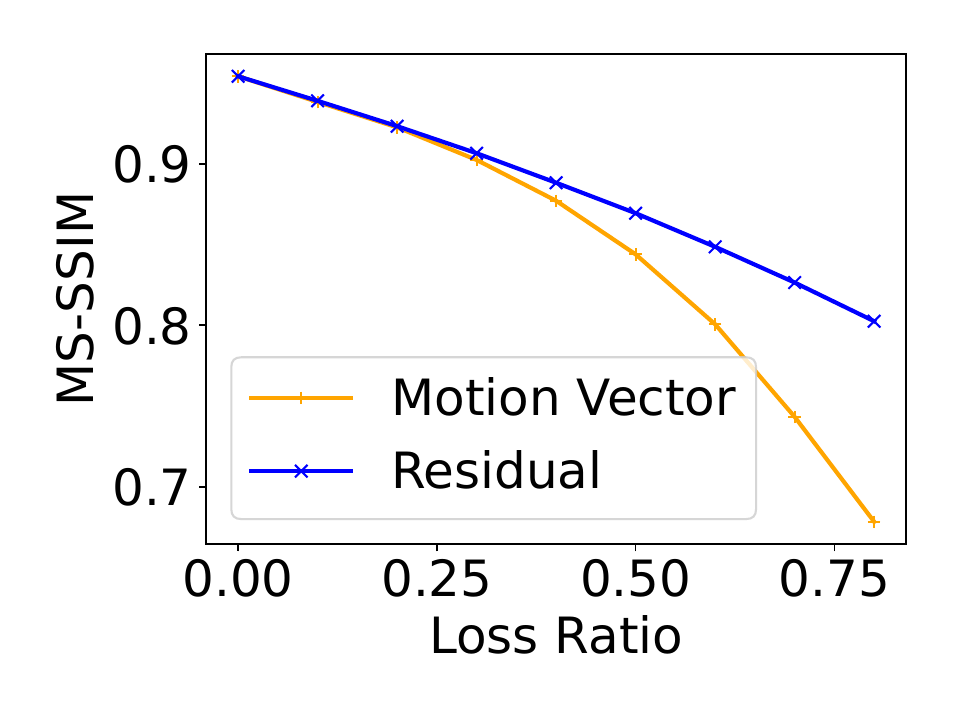}}
    \vspace{-4mm}
    \caption{Impact of losses of motion vs. residuals on the video quality of Grace~\cite{cheng2024grace}, a loss-resilient residual-coding codec. Figure labels indicate which source information is corrupted while the other is fully received. }
    \vspace{-6mm}
    \label{f:source_type_loss_perf}
\end{minipage}
\end{figure}


\subsection{\textbf{Source Information Splitting}}
The source information considered by \algorithmname codec is the latent representation of each frame. 
The representation generated by AutoEncoder transform exhibits spatial-temporal correlations~\cite{li2023mage, yu2023magvit} and an information compaction property~\cite{he2022elic}: a few channels exhibit significantly higher average energy (see channel map visualization example in \fig \ref{fig:channel_maps}). 
Since channels with higher energy are more important, we split the latent representation by masking out portions of channel maps to ensure resilience to description loss.  Instead of treating each of the $h \times w \times c$ elements in the representation as a token, we group each $1 \times 1 \times c$ column into a token and split these $hw$ tokens into different descriptions. This approach maintains a similar energy level in each description and avoids creating infeasibly long sequences for transformers. \reff{seperation_example } shows an example of forming 4 descriptions. We split the latent representation by masking out it with a special learnable mask token in an interleaving way\footnote{Random splitting also works as long as it is reversible at the receiver side. We use interleaving splitting for its simplicity and similar performance to random splitting.}, forming multiple masked latent representations whose combination equals the original representation.

Note that we do not utilize other types of source information, such as motion, optical flow, or residuals~\cite{lu2019dvc,xiang2022mimt,li2023neural}, as they carry differently important source information and lack strong correlations with each other. Consequently, the loss of one type (\eg motion) cannot be efficiently estimated from the received other types (\eg residual). 
The distinct impacts of loss on motion vectors and residuals on reconstructed video quality are shown in \fig \ref{f:source_type_loss_perf}. 
It is evident that motion information is more critical than residuals, and the loss of motion cannot be effectively compensated for, even if the residuals are fully received.  
Therefore, our \algorithmname video codec exclusively uses the latent representation as source information, letting the transformer extract diverse contexts from representations for compression.

\begin{figure}[t]
\vspace{-2mm}
    \centering
    \vspace{-4mm}
    \includegraphics[width=\linewidth]{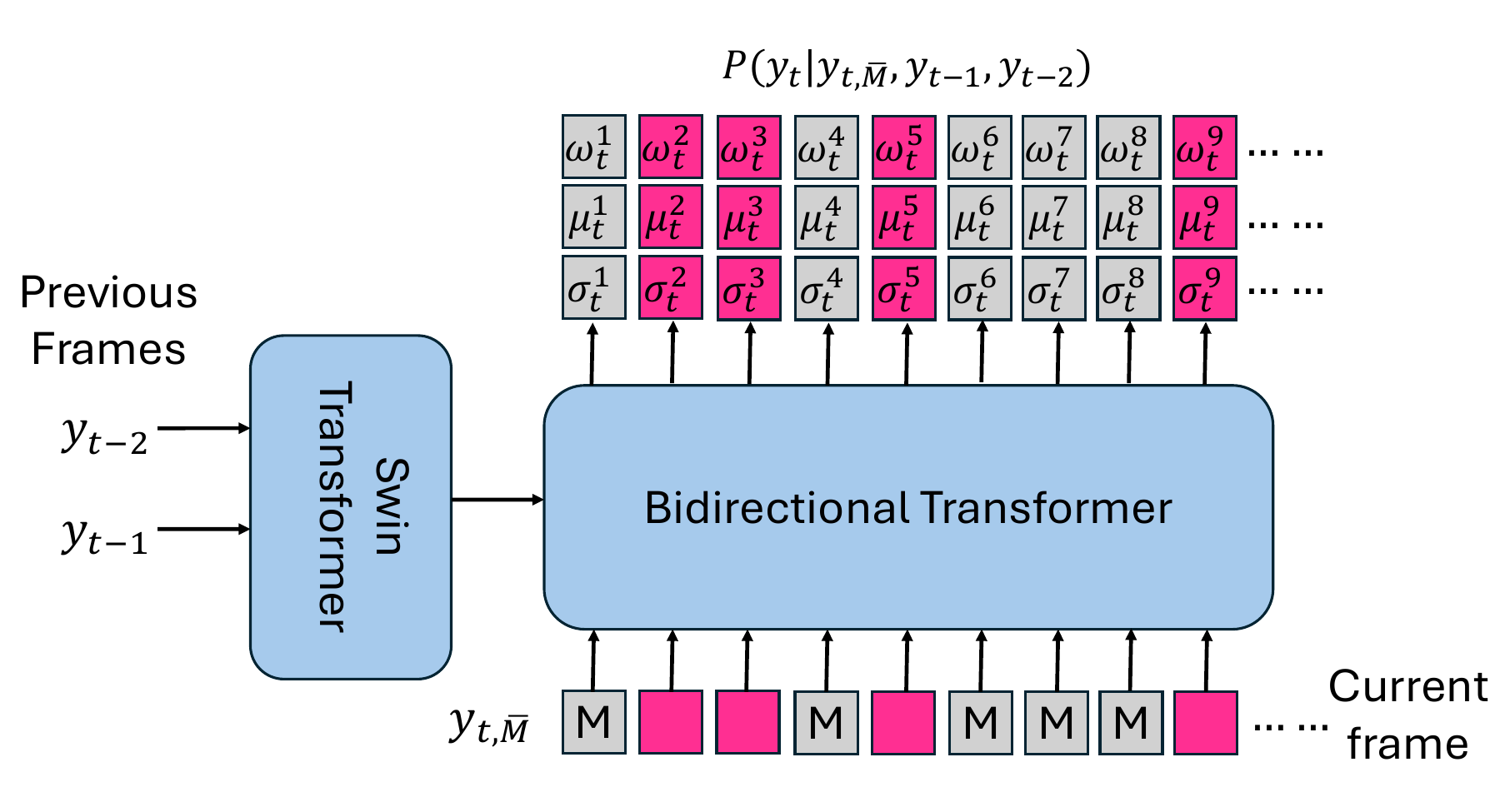}
    \vspace{-6mm}
    \caption{Overview of the masked transformer entropy model. During training, the model learns to predict the distributions of masked tokens. At inference, the model begins by predicting all masked tokens and then follows the QLDS masking schedule to keep a portion of predicted tokens as input for the next prediction iteration. This process continues until all tokens are uncovered.}
   \vspace{-4mm}
    \label{f:entropy_model}
\end{figure}

\subsection{\textbf{Masked Spatial-Temporal Transformer Entropy Coding}}

We independently entropy encode each description into an MDC stream, allowing each stream to be decoded independently. To reduce the bit length of each stream, we propose a masked spatial-temporal transformer entropy model. 
Given a sequence of video frames $\{x_t\}_{t=1}^T$ and the corresponding latent representation sequence $\{y_t\}_{t=1}^T$, where each representation is split into $S$ descriptions $y_t = [y_{t,s}]_{s-1}^S$, we use the masked transformer to predict  $P\{y_{t,s} | y_{t-1}, y_{t-2}\}$ and entropy code each description $y_{t,s}$ to a bitstream.  The transformer runs independently on each description, trading reduced spatial context for parallel execution. To compress the full video, we simply apply this procedure iteratively, letting the transformer predict the conditional distribution for each latent representation and padding with zeros when predicting distributions for the first two frames.

\paragraph{Entropy Model } \reff{entropy_model} shows the overview of our masked transformer entropy model. It extends MaskGIT-like transformer~\cite{chang2022maskgit, m2t} to extract both spatial and temporal dependencies among latent tokens. 
Let $y_t = [y_t^i]_{i=1}^N$ denote the latent tokens of the current frame, where $N$ is the length of the reshaped token matrix, and M$=[m_i]_{i=1}^N$ is the corresponding binary mask. To simulate an arbitrary number of descriptions during training, we randomly sample a subset (from $0\%$ to $100\%$) of tokens and replace them with a special learnable mask token $[M]$. $m_i=1$ indicates that the token $y_t^i$ is replaced with  $[M]$. Denote $y_{t, \overline{M}}$ as the masked representation after applying mask M to $y_t$. We train the masked transformer to minimize the cross entropy of the predicted distributions $P$ with respect to the true distribution $Q$, \ie the average bit rate:
\begin{equation}
    R(y_t) = E_{y_t \sim Q}[
    \sum_{m_i=1} -\log_2 p(y_t^i |y_{t, \overline{M}}, y_{t-1}, y_{t-2})]
\end{equation}
Here, previous representations  $y_{t-1}$,  $y_{t-2}$, and the current masked representation $y_{t, \overline{M}}$ provide both temporal and spatial context for predicting the distribution of masked tokens.
We model this conditional distribution through a mixture of Gaussians (GMM) with $N_M=3$ mixtures, each parameterized by a mean $\mu$, scale $\sigma$, and weight $\omega$. 

\paragraph{Iterative Encoding and Decoding} 
One approach is to use the above masked transformer entropy model to encode and decode a description $y_{t,s}$ in one step by masking out all latent tokens in the description. However, this is inefficient because the spatial context information in the description is totally ignored and hence increases the bitrate cost. Instead, we apply the entropy model $L$ times following the QLDS masking schedule~\cite{m2t} $\{\text{M}_1, ..., \text{M}_L\}$, where $\text{M}_i[j]=1$ indicates that the j-th token is predicted and uncovered at step $i$ and the number of tokens uncovered monotonically increases during iteration. 
At the first iteration, we start with all tokens in $y_{t,s}$ are $[M]$, then we only entropy code the tokens corresponding to $\text{M}_1$, uncover them as input for the next iteration. The process repeats until all tokens in $y_{t,s}$ have been entropy coded and uncovered. 

Note that, unlike VCT~\cite{2022-VCT}, which uses transformers to model the distribution autoregressively and sequentially, the masked bi-directional transformer predicts the distribution with richer contexts by attending to all tokens in the provided representations. To mitigate the impact of temporal error propagation caused by corrupted previous frames, in contrast to MIMT~\cite{xiang2022mimt}, \algorithmname utilizes only the two most correlated latent representations from the past. Furthermore, to ensure each description is independently decodable and robust to potential loss, \algorithmname avoids conditioning on any side information, such as hyper-prior and optical flow used by MIMT, since their loss cannot be efficiently estimated.

\subsection{Loss and Training Process}
We decompose the training into three stages. In \textbf{stage I}, we train the per-frame encoder and decoder by minimizing the rate-distortion trade-off $r(y) + \lambda d(x, \widetilde x)$:
\begin{equation}
    L_I = E_{x\sim p_X, \mu \sim U \pm0.5} [-\log p(\hat{y} + \mu) + \lambda MSE(x, \hat{x})]
    \label{eq:rd}
\end{equation}
where $x\sim p_X$ are frames drawn from the training set, $\hat{y}$ refers to the unquantized representation, and we use additive i.i.d. noise from a uniform distriubiton in $[-0.5, 0.5]$ to simulate quantization during training~\cite{theis2022lossy}. We use mean squared error (MSE) as the distortion loss and employ the mean-scale hyperprior~\cite{minnen2018joint} approach to estimate $p$ (\ie bitrate) temporally, which we discard in later stages. To get gradients through the quantization operation, we rely on straight-through estimation (STE)~\cite{minnen2020channel, theis2022lossy}. After stage I, we obtain the lossy ELIC encoder and decoder transformers reaching nearly any desired distortion $MSE(x, \hat{x})$ by varying how large the range of each element in $y$ is. Basically, the wider the value range of $y$, the higher the quality of frame reconstruction and the larger the bitrate tends to be.

In \textbf{stage II}, we train the masked temporal transformer to obtain $p$ and only minimize the bitrate:
\begin{multline}
    L_{II} = E_{(x_1, x_2, x_3)\sim p_{X_{1:3}}, \mu \sim U}[\\
    \sum_{M[i]=1} -\log_2 p(y_{3}^i + \mu | y_{3,\overline{M}}, y_1, y_2)]
    \label{eq:bitrate}
\end{multline}
where $(x_1, x_2, x_3)\sim p_{X_{1:3}}$ are three adjacent video frames. 
Given the representation $y$, we randomly sample a mask $M$, where 0-100\% of the entries are 1. The corresponding entries in $y$ are masked, which means we replace them with a special mask token, which is a learned $c$-dimensional vector. Together with the previous representation $y_1$,  $y_2$ , the resulting masked representation $y_{3,\overline{M}}$, which simulates the description after any arbitrary source splitting, are fed to the masked temporal transformer, which predicts the distribution of the tokens. We assume the distribution of each token is a mixture of Gaussian and let the transformer predict the mean, scale, and weight per token. 
When computing the bitrate loss, we only consider the distributions corresponding to the masked tokens.

In \textbf{stage III}, we finetune the ELIC Autoencoder and the masked transformer jointly by replacing the mean-sale hyperprior entropy model with the masked transformer entropy model (\ie replacing $p()$ in Eq.~\ref{eq:rd} with  Eq.~\ref{eq:bitrate}).


\subsection{Inference of Lost Tokens}
The inference of lost latent tokens involves predicting and sampling. 
After entropy decoding the received streams and merging available tokens, to predict the lost tokens caused by transmission loss, we apply the masked transformer to predict the probabilities, denoted as $p(y_{t, M} |\widetilde y_{t,\overline M}, \widetilde y_{t-1}, \widetilde y_{t-2})$, for all the masked tokens in parallel. Here, the reconstructed tokens of current frame $\widetilde y_{t,\overline M}$ and previous representations provide temporal and spatial contexts for the transformer to predict the distributions of the missing tokens. At each masked location $j$, we sample a token $y_{t}^j$ based on its maximal probabilities
\begin{equation}
    \widetilde y_{t}^j = \argmax_{y_{t}^j} p(y_{t, M} |\widetilde y_{t,\overline M}, \widetilde y_{t-1}, \widetilde y_{t-2})
\end{equation}

\section{Experiments} \label{s:experiment}


    
    

\paragraph{Datasets.} We train \algorithmname on the Vimeo-90K dataset~\cite{xue2019video}. During training, we randomly sample 256 $\times$ 256 crops from the original frames. The training batches are made up of randomly selected triplets of adjacent frames.  
We evaluate on two common benchmark datasets: UVG~\cite{mercat2020uvg} and MCL-JCV~\cite{wang2016mcl}, both containing raw videos with a resolution of $1920\times1080$. 

\begin{figure}
    \centering
    \hspace*{-.22in}\includegraphics[width=1.1\linewidth]{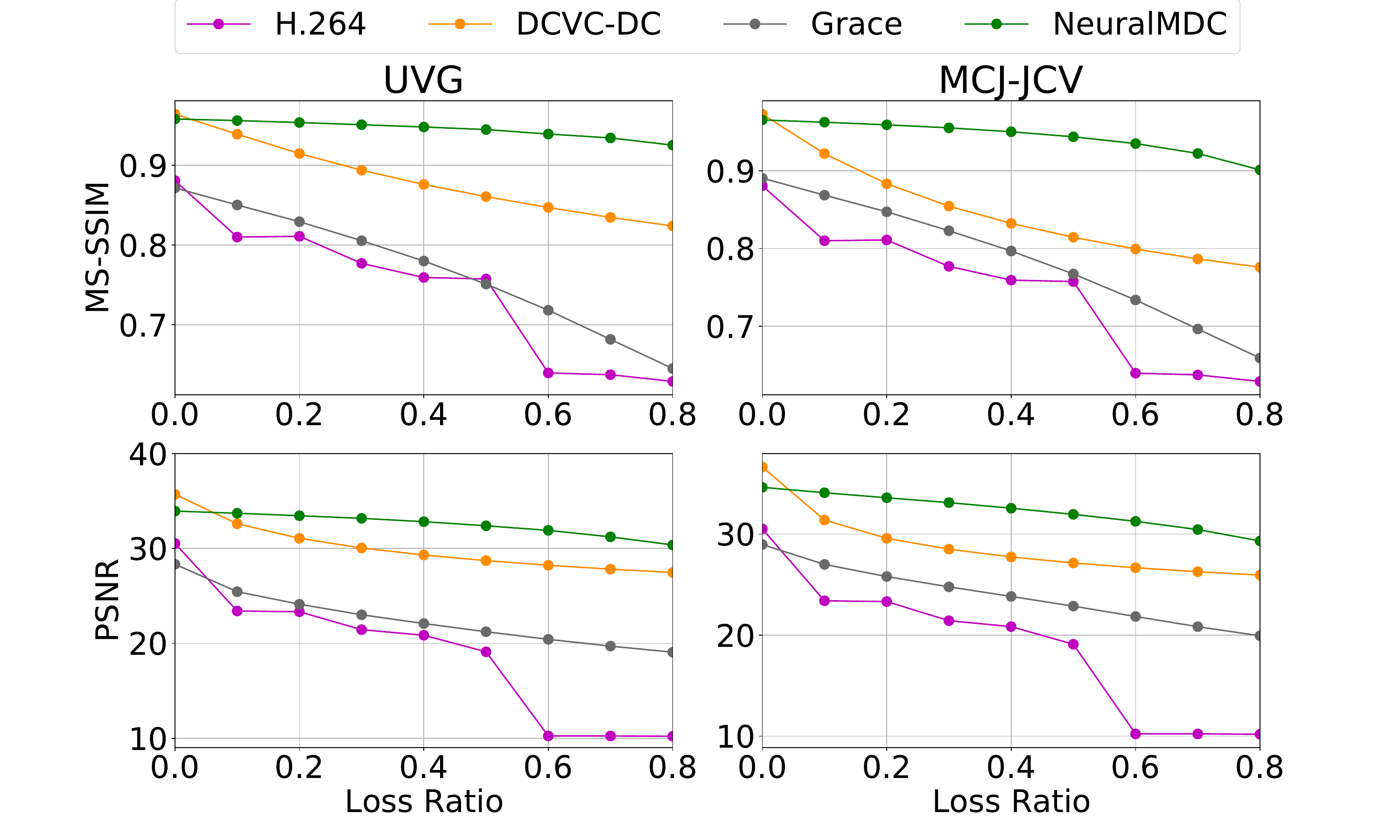}
    \vspace{-6mm}
    \caption{At the same bitrate and without retransmission, reconstructed video quality achieved by different codecs under varying loss rates. } 
    \label{f:loss_resiliency}
    \vspace{-4mm}
\end{figure}

\paragraph{Baselines.} 
We compare \algorithmname against the following video codecs for evaluating loss resilience and rate-distortion performance. The implementation details of \algorithmname are elaborated in the Appendix \ref{a:implementation}. 
For loss resilience evaluation, we randomly corrupt bitstreams of \textbf{H.264} using the FFmpeg x264 codec based on the bitstream corruption framework in ~\cite{liu2024bitstream}. We run \textbf{Grace}~\cite{cheng2024grace}, a loss-resilient residual-coding video codec, using their public checkpoints. Grace is an extension of DVC and trains a variational autoencoder where the latent representation is sampled from a specific loss distribution. We also run \textbf{DCVC-DC}~\cite{li2023neural} using their public checkpoints to evaluate the loss resilience of the condition coding-based codec. 
As for the rate-distortion performance, we further obtain reported results from the following papers: \textbf{VCT}~\cite{2022-VCT}, \textbf{C2F}~\cite{hu2022coarse}, \textbf{ELF-VC}~\cite{rippel2021elf}, \textbf{DCVC}~\cite{li2021deep}, \textbf{FVC}~\cite{2021-FVC}, \textbf{DVC}~\cite{lu2019dvc}. All experiments are conducted with one AMD Ryzen Threadripper PRO 3995WX 64-Cores CPU and one Nvidia RTX A6000 GPU. 


\paragraph{Metrics.} We evaluate the common visual quality metrics, PSNR and MS-SSIM~\cite{wang2003multiscale} in RGB.

\section{Results}

\subsection{Loss Resilience Performance}

In the real world, video streaming over communication networks can experience packet loss (referring to both packets dropped\footnote{See Appendix \reff{dynamic_traces} for packet drop rates of a real-world 5G trace example. } in transit and those not received by the decoding deadline) ranging from 0\% to over 80\%~\cite{cheng2024grace}.
In this section, we  compare \algorithmname with other video codecs that operate without retransmission and with DCVC-DC, which operates with retransmission.
\begin{figure}
    \hspace*{-.15 in}\centering{\includegraphics[width=1.1\linewidth]{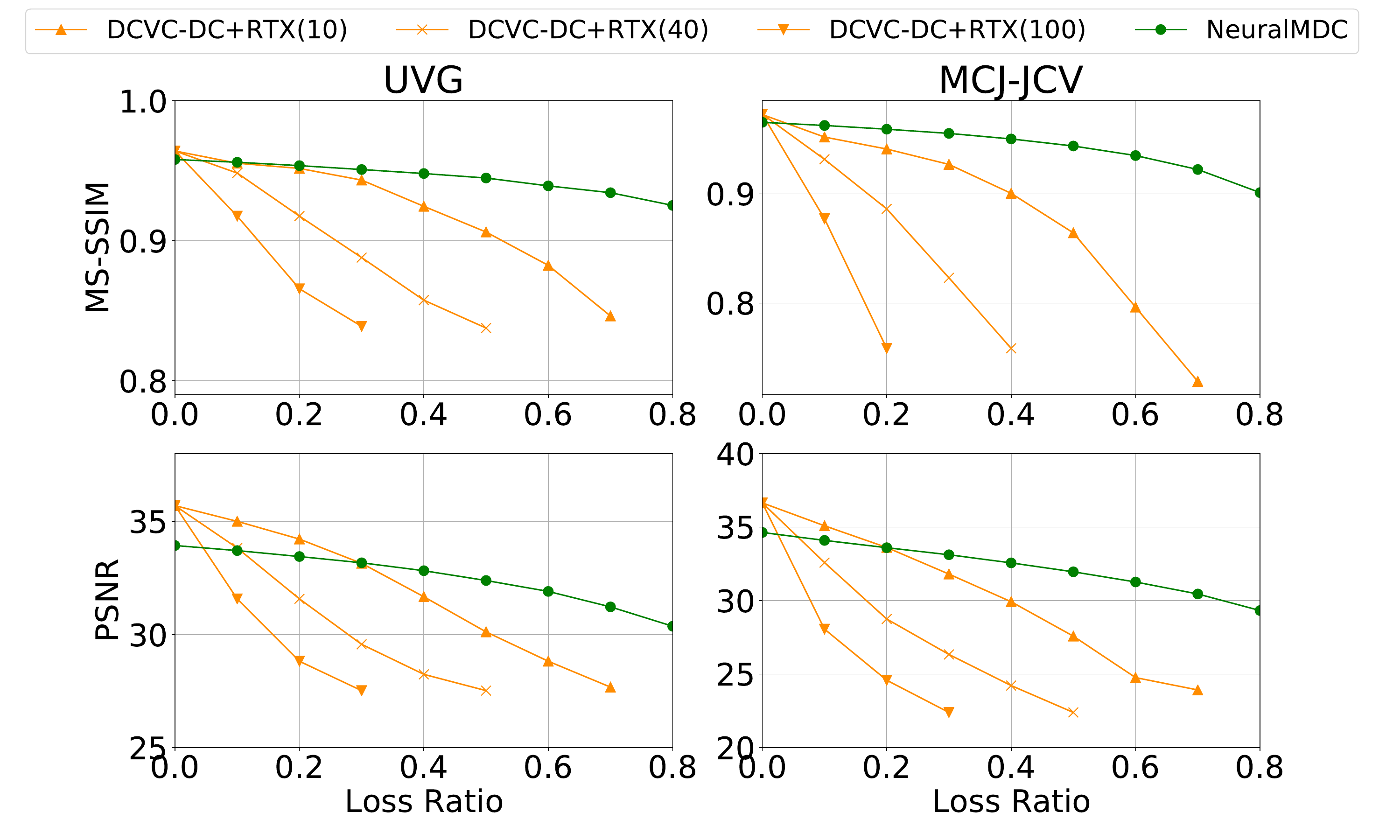}}
    \vspace{-6mm}
    \caption{Comparison between NeuralMDC without retransmission and DCVC-DC with RTX($t$) across various network round-trip time($t$ ms)  and loss rates under the same network bandwidth and transmission time.} 
    \label{f:loss_resiliency_rtx}
    \vspace{-6mm}
\end{figure}

\begin{figure*}
    \vspace{-.2cm}
    \hfill\subfigure{\includegraphics[width=.24\linewidth]{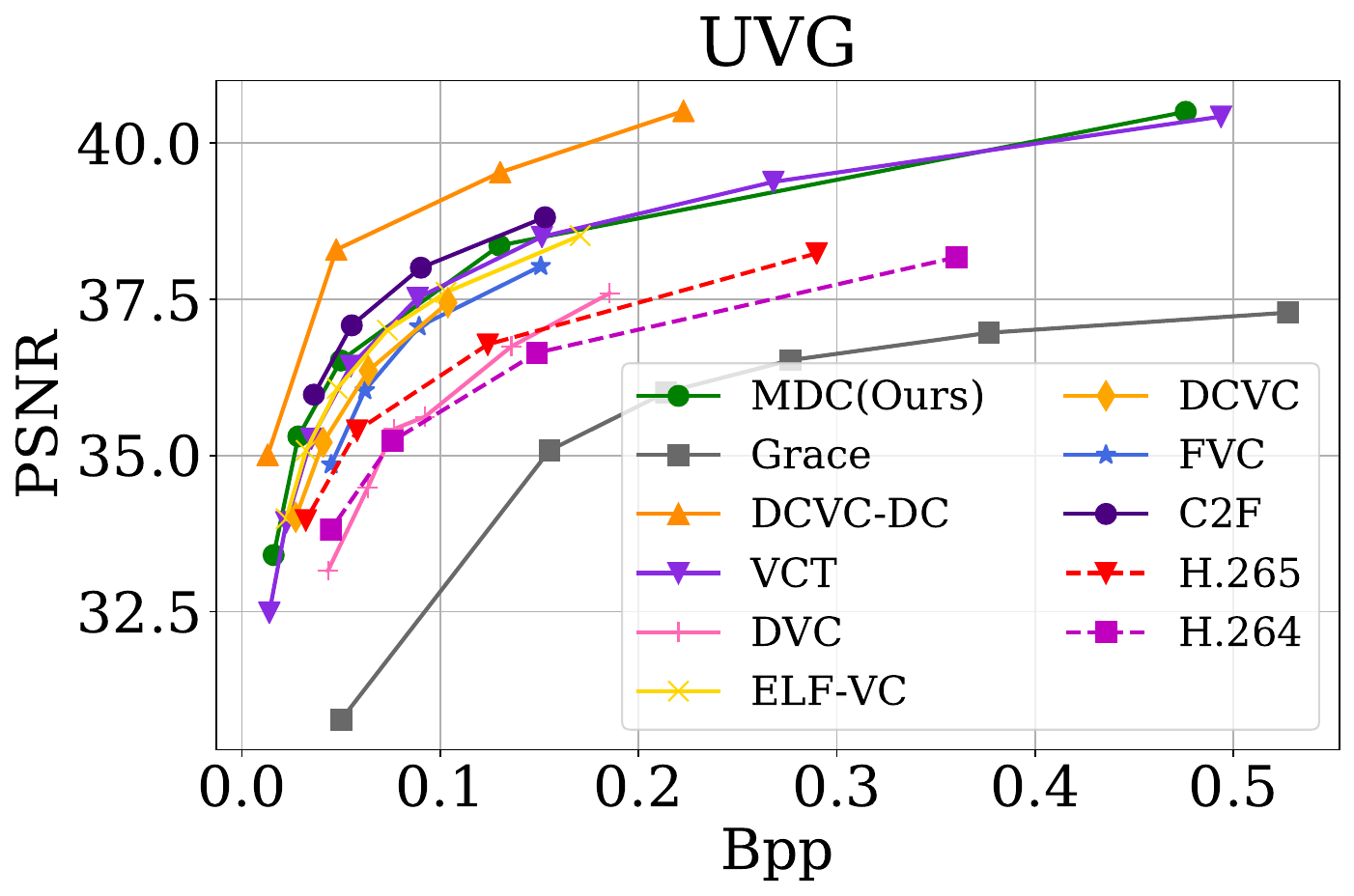}}
    \hfill\subfigure{\includegraphics[width=.24\linewidth]{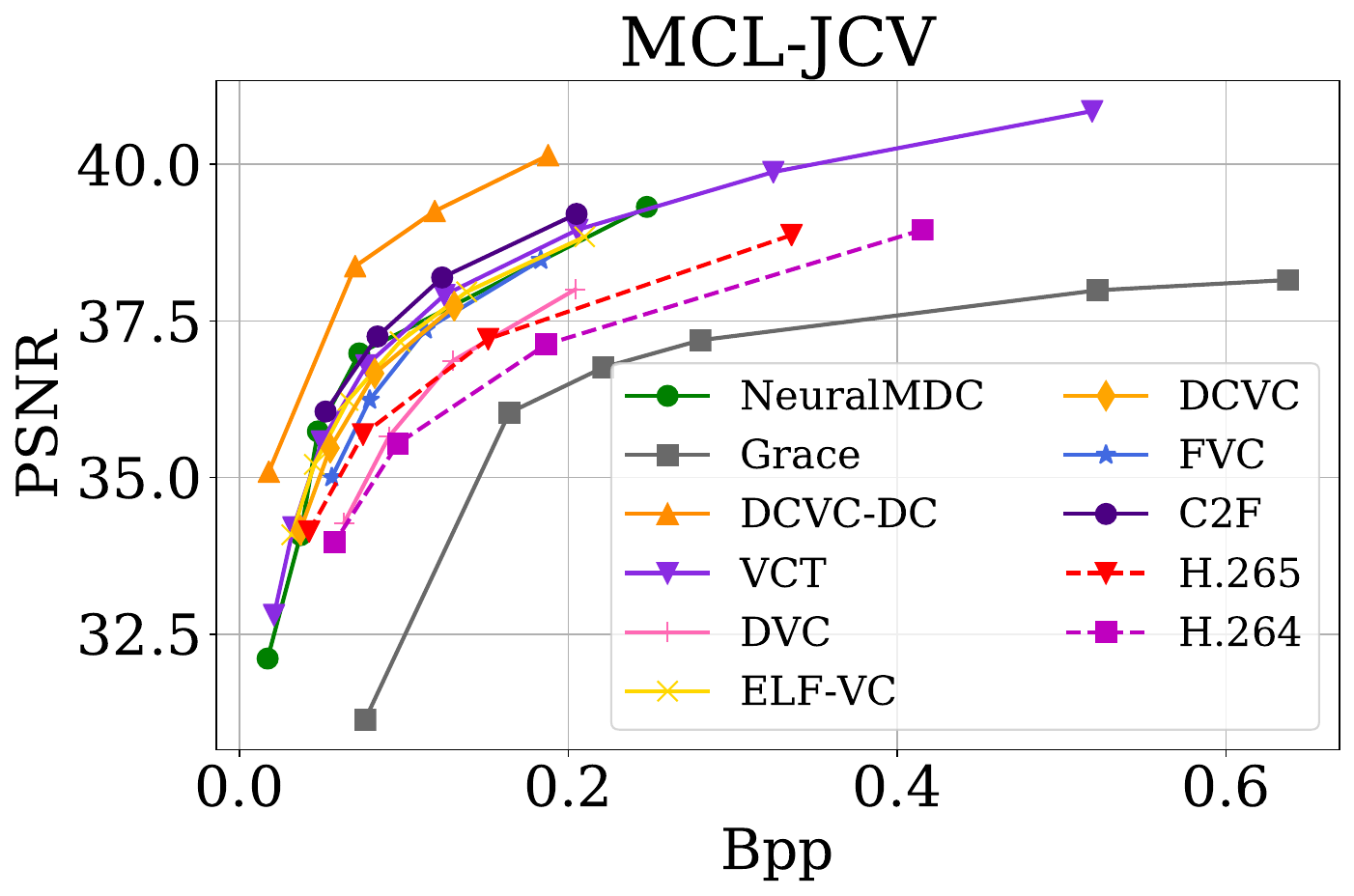}}
    \subfigure{\includegraphics[width=.24\linewidth]{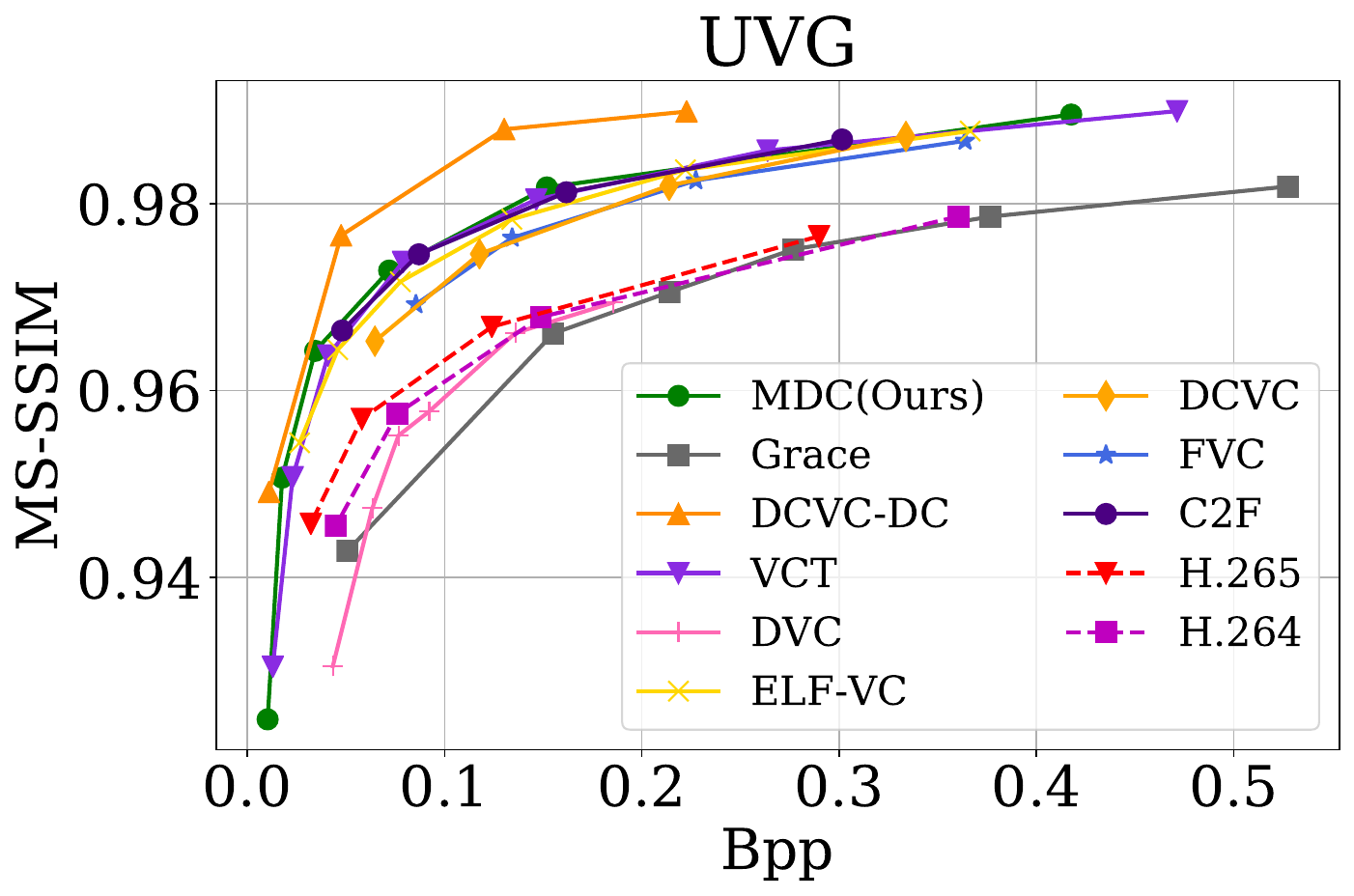}}
    \hfill\subfigure{\includegraphics[width=.24\linewidth]{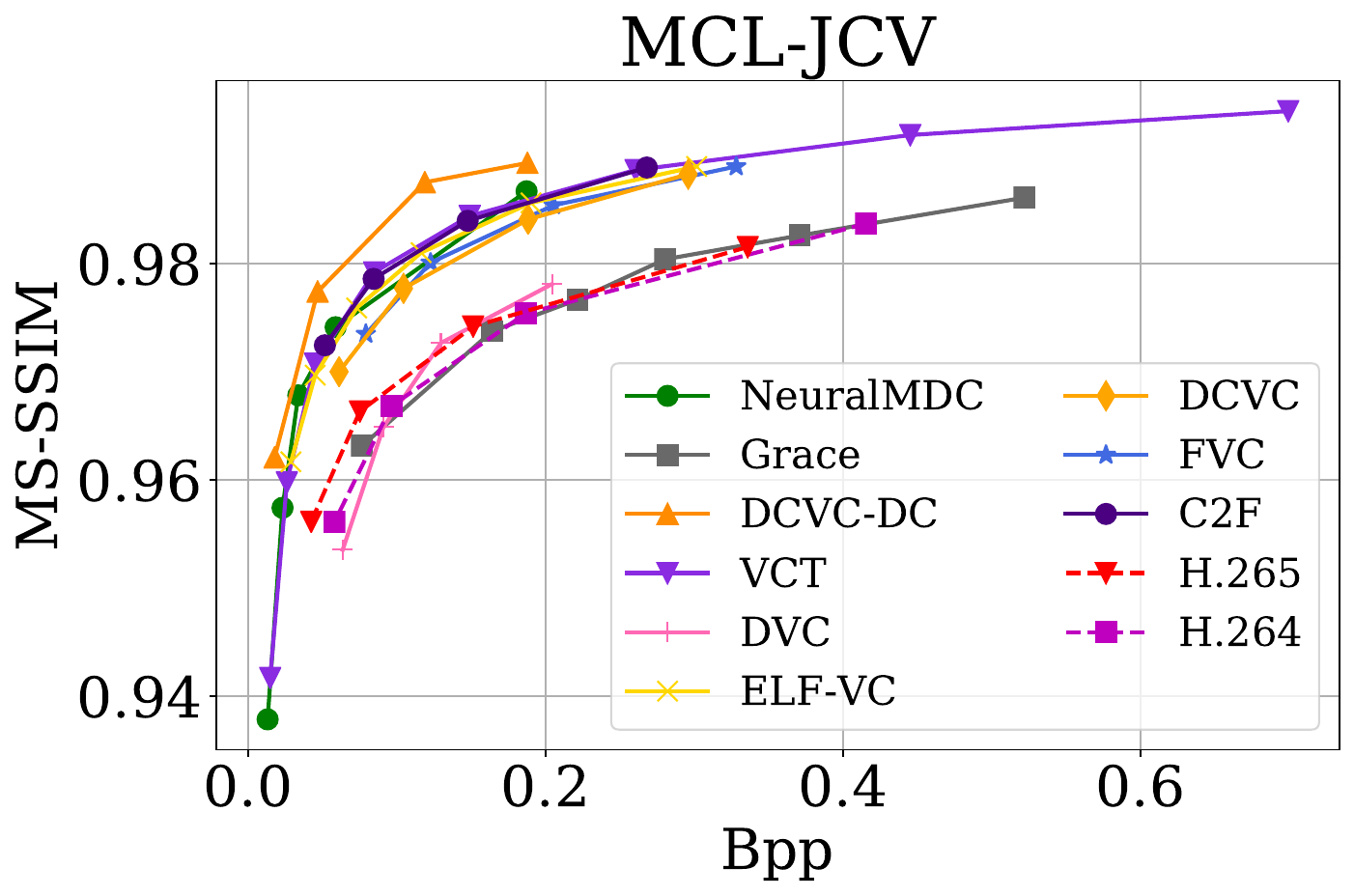}}
    \vspace{-.4cm}
    \caption{Rate-distortion performance on UVG and MCL-JCV datasets.}
    \vspace{-.2cm}
    \label{f:compression_efficiency}
\end{figure*}

\paragraph{Baselines without Retransimission}
\reff{loss_resiliency} compares the decoded video quality of \algorithmname with the baselines under varying loss rates on the UVG and MCL-JCV datasets. For a fair comparison, we ensure that \algorithmname and all baselines have the similar bpp performance and none of them retransmit lost packets.  
We see that our \algorithmname outperforms all the baselines in both PSNR and MS-SSIM. The loss resilience performance of our \algorithmname surpasses the best baseline by 1.78 to 8.66 times. 
Although DCVC-DC achieves higher visual quality in the absence of packet loss, it is highly sensitive to packet loss, causing the reconstructed video quality to degrade more rapidly compared to \algorithmname. 
This verifies the effectiveness of our masked transformer entropy model in exploiting the spatial and temporal redundancies in received descriptions to infer the lost tokens. 
Since both Grace and DCVC-DC utilize motion information and DCVC-DC propagates extracted features along frames, their poor performance indicates that lost motion cannot be efficiently estimated and the temporal error caused by encoder-decoder state mismatch propagates due to feature propagation. 
The visualization of reconstruction samples under $50\%$ loss rate is shown in Appendix \reff{reconstruct}. 

\begin{figure*}[t]
\begin{minipage}{.31\textwidth}   
    \centering
    \includegraphics[width=.9\linewidth]{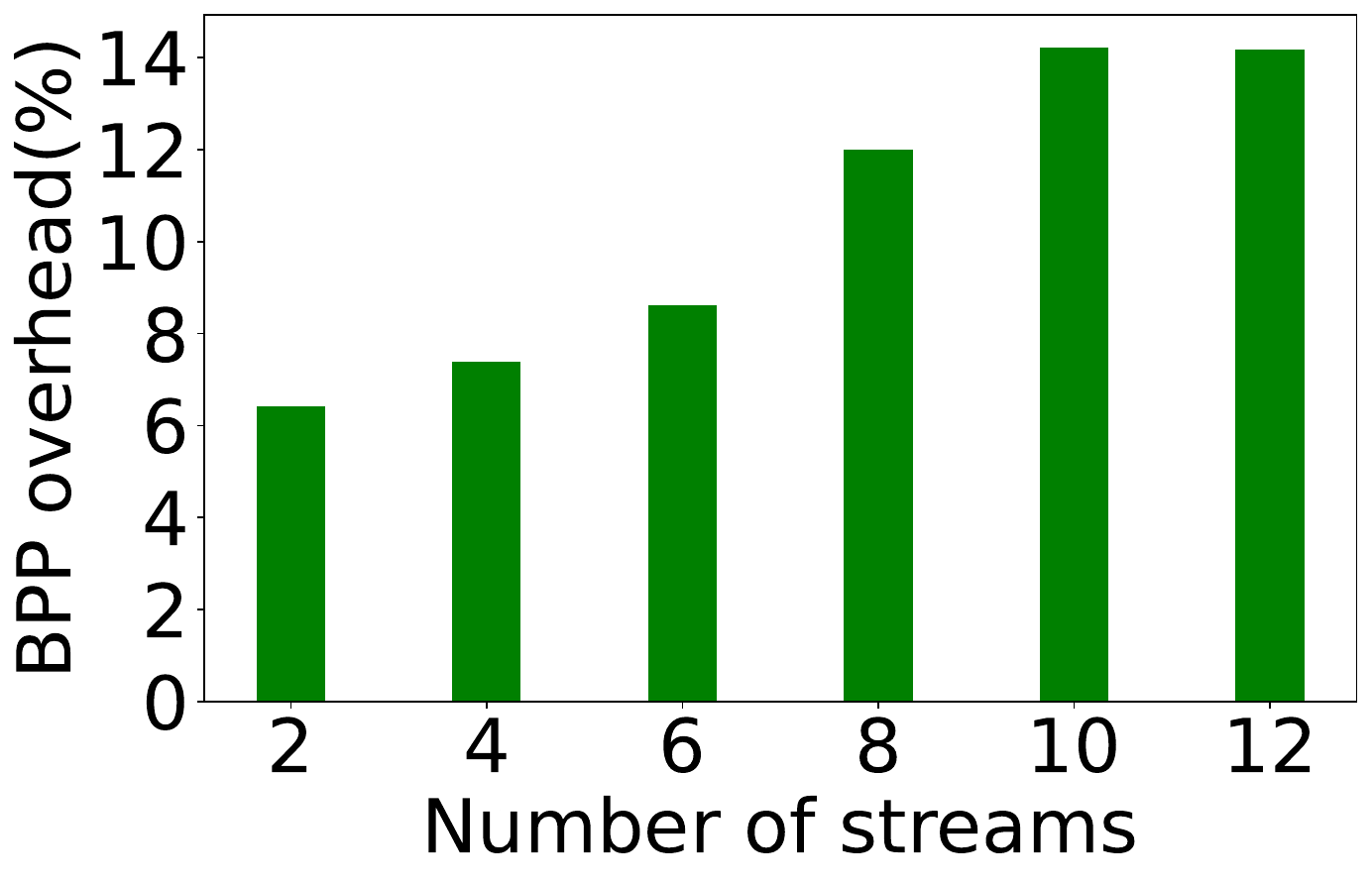}
    \vspace{-2mm}
    \caption{The average BPP overhead incurred by multiple descriptions.}
    \label{f:mdc_bpp_overhead}
    \end{minipage}
\hfill
\begin{minipage}{.31\textwidth}   
    \centering
    \includegraphics[width=.9\linewidth]{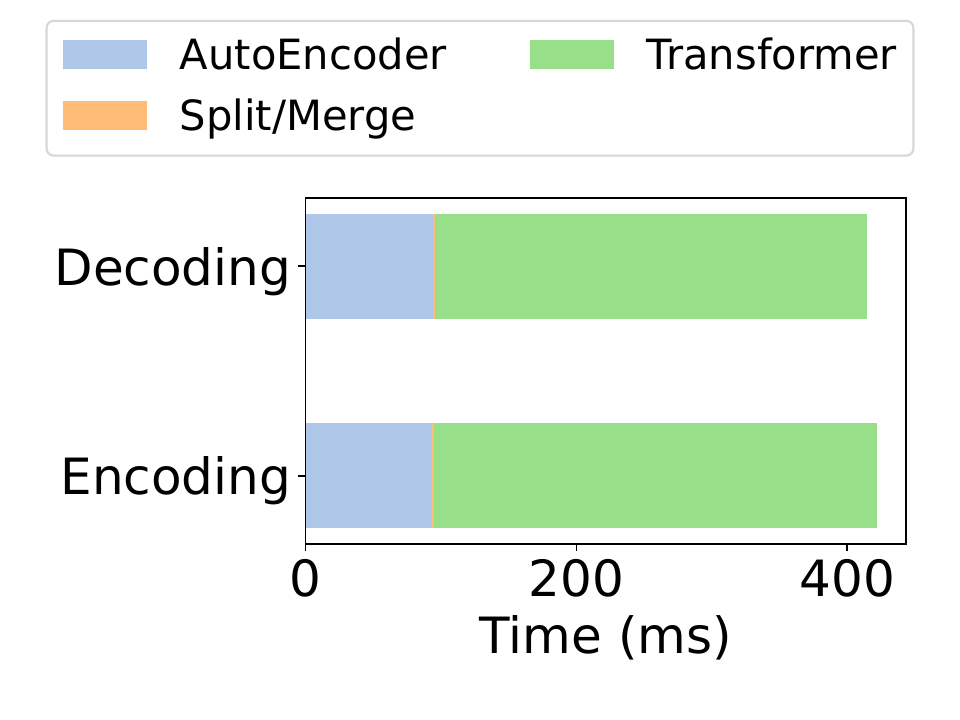}
    \vspace{-6mm}
    \caption{Runtime breakdown of \algorithmname on a 1080p frame.}
    \label{f:runtime}
    \end{minipage}
\hfill
\begin{minipage}{.31\textwidth}
    \centering
    \includegraphics[width=0.9\linewidth]{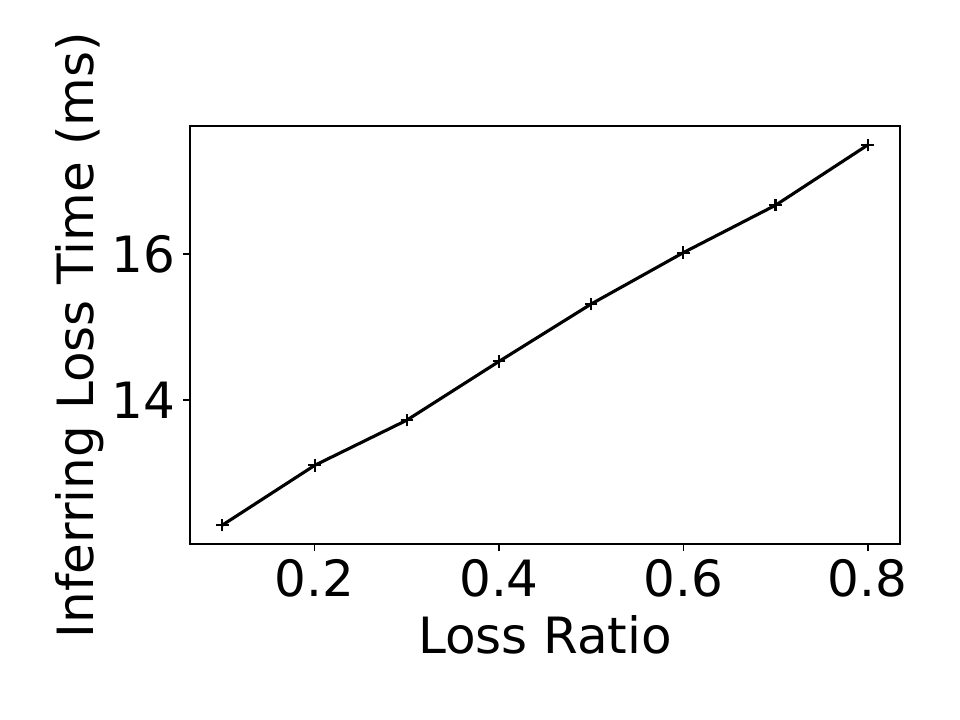}
    \vspace{-6mm}
    \caption{Runtime of inferring lost tokens on a 1080p frame.}
    \label{f:runtime_infer}
    \end{minipage}
    \vspace{-4mm}
\end{figure*}

\paragraph{Baselines with Retransimission} 
Since DCVC-DC has a better rate-distortion performance than \algorithmname, we further compare \algorithmname with DCVC-DV that  operates additionally with the retransmission (RTX) scheme\footnote{Here the retransmission scheme refers to both the retransmission of packets dropped in transit and the reinjection of packets from low paths to fast paths~\cite{zheng2021xlink}.}. In this experiment, both \algorithmname  and DCVC-DC are evaluated under the same network bandwidth and transmission time. This means that as the loss ratio increases, the
effective bpp of DCVC-DC with RTX decreases. We also consider the impact of network round-trip time (RTT).
Typically, modern transport protocols wait 1.5 times the RTT to retransmit a lost packet~\cite{stevens1997rfc2001}. Consequently, to transmit videos within the same time, a higher RTT  further reduces the effective bpp of DCVC-DC with RTX. \reff{loss_resiliency_rtx} shows that \algorithmname outperforms DCVC-DC with RTX when RTT exceeds 10 ms. As RTT increases, the loss resilience performance advantage of NeuralMDC over DCVC-DC further improves. This highlights the superior performance of  \algorithmname in achieving high visual quality and low latency video streaming, even when compared to state-of-the-art neural codecs protected by the RTX scheme.

\subsection{Rate-Distortion Performance}
\reff{compression_efficiency}  shows the rate-distortion performance in scenarios where no packet loss occurs. 
In this case, we set the number of descriptions to 1 and evaluate the overhead of source splitting later. 
Except for DCVC-DC, our \algorithmname outperforms VCT and other baselines. This demonstrates the effectiveness of the bidirectional transformer in extracting richer contexts to improve compression efficiency. 
Our \algorithmname has a worse rate-distortion trade-off compared to DCVC-DC because DCVC-DC additionally utilizes motion vectors to extract more contexts and propagates extracted features along frames. However, as shown above, this makes DCVC-DC more sensitive to packet loss.
We note that Grace sacrifices a significant amount of compression efficiency to make DVC robust to packet loss.

\subsection{\algorithmname BPP Overhead}
Since \algorithmname splits the source information into distinct descriptions, the bit costs increase because the correlation of source information within each description decreases, thereby reducing compression efficiency. 
\reff{mdc_bpp_overhead} shows the bpp overhead caused by source splitting with respect to the anchor of  a single description. 
As the number of descriptions increases, the bpp overhead also increases. However, our \algorithmname codec exhibits an upper limit on the bpp overhead increase. This is because, as the number of descriptions grows, the previous frame information primarily provides the context for compression, even though the decreased correlation within the current frame information offers little context.


\subsection{Runtime}

We conduct a detailed breakdown of the time costs associated with \algorithmname. The video codec is tested with 1080p videos. The masked transformer entropy model is applied 12 times to iteratively encode and decode each description, following the QLDS masking schedules. We run the transformer in parallel for 4 descriptions.  \reff{runtime} shows the runtime of the encoding and decoding processes. Note that running the transformer at 1080p once only takes about 27.29 ms, but we run it 12 times for iteratively entropy encoding and decoding. 

\reff{runtime_infer}  shows the inference time for predicting lost tokens due to packet loss. We only report the runtime of inferring tokens from the predicted representation distributions, as we can reuse the distribution prediction results in the entropy decoding stage. As the packet loss ratio increases, the inference time also increases. This is reasonable because higher loss means more tokens needs to be inferred from the predicted distribution, which requires more computational resources and time.


\section{Conclusions} \label{s:conclusion}
We have designed a novel error-resilient source coding method, \algorithmname, for video delivery over dynamic and noisy networks. It is designed in particular to take advantage of dynamically available, albeit noisy, multiple network paths or radio channels that have become prevalent in today's high-speed networks such as 5G. 
\algorithmname first tokenizes each input frame into its latent representation.
It then splits the tokens on the channel-axis to evenly distribute energy among different descriptions for redundancy allocation.
\algorithmname finally trains a spatial-temporal masked transformer to capture the spatial-temporal correlations of tokens.
Furthermore, \algorithmname performs token entropy coding based on
distributions derived from the trained transformer to achieve efficient compression.
For error-resilient decoding, \algorithmname infers missing tokens using received current and past tokens and reconstructs frames using both received and inferred tokens.
We show that \algorithmname exhibits a superior 2 to 8 times improvement in loss resilience while achieving compression efficiency comparable to the state-of-the-art.

\noindent

\noindent


\clearpage

\bibliography{bib/reference}  

\begin{thebibliography}{53}
\providecommand{\natexlab}[1]{#1}

\bibitem[{Agustsson et~al.(2020)Agustsson, Minnen, Johnston, Balle, Hwang, and Toderici}]{2020-scale}
Agustsson, E.; Minnen, D.; Johnston, N.; Balle, J.; Hwang, S.~J.; and Toderici, G. 2020.
\newblock Scale-space flow for end-to-end optimized video compression.
\newblock In \emph{Proceedings of the IEEE/CVF Conference on Computer Vision and Pattern Recognition}, 8503--8512.

\bibitem[{Badr et~al.(2017)Badr, Khisti, Tan, Zhu, and Apostolopoulos}]{badr2017fec}
Badr, A.; Khisti, A.; Tan, W.-t.; Zhu, X.; and Apostolopoulos, J. 2017.
\newblock FEC for VoIP using dual-delay streaming codes.
\newblock In \emph{IEEE INFOCOM 2017-IEEE Conference on Computer Communications}, 1--9. IEEE.

\bibitem[{Chang et~al.(2022)Chang, Zhang, Jiang, Liu, and Freeman}]{chang2022maskgit}
Chang, H.; Zhang, H.; Jiang, L.; Liu, C.; and Freeman, W.~T. 2022.
\newblock Maskgit: Masked generative image transformer.
\newblock In \emph{Proceedings of the IEEE/CVF Conference on Computer Vision and Pattern Recognition}, 11315--11325.

\bibitem[{Chen et~al.(2021)Chen, He, Wang, Ren, Lim, and Shrivastava}]{chen2021nerv}
Chen, H.; He, B.; Wang, H.; Ren, Y.; Lim, S.~N.; and Shrivastava, A. 2021.
\newblock Nerv: Neural representations for videos.
\newblock \emph{Advances in Neural Information Processing Systems}, 34: 21557--21568.

\bibitem[{Cheng et~al.(2024)Cheng, Zhang, Li, Arapin, Zhang, Zhang, Liu, Du, Zhang, Yan et~al.}]{cheng2024grace}
Cheng, Y.; Zhang, Z.; Li, H.; Arapin, A.; Zhang, Y.; Zhang, Q.; Liu, Y.; Du, K.; Zhang, X.; Yan, F.~Y.; et~al. 2024.
\newblock $\{$GRACE$\}$:$\{$Loss-Resilient$\}$$\{$Real-Time$\}$ Video through Neural Codecs.
\newblock In \emph{21st USENIX Symposium on Networked Systems Design and Implementation (NSDI 24)}, 509--531.

\bibitem[{Conci and De~Natale(2007)}]{conci2007real}
Conci, N.; and De~Natale, F.~G. 2007.
\newblock Real-time multiple description intra-coding by sorting and interpolation of coefficients.
\newblock \emph{Signal, Image and Video Processing}, 1: 1--10.

\bibitem[{Fleming and Effros(1999)}]{fleming1999generalized}
Fleming, M.; and Effros, M. 1999.
\newblock Generalized multiple description vector quantization.
\newblock In \emph{Proceedings DCC'99 Data Compression Conference (Cat. No. PR00096)}, 3--12. IEEE.

\bibitem[{Franchi et~al.(2005)Franchi, Fumagalli, Lancini, and Tubaro}]{franchi2005multiple}
Franchi, N.; Fumagalli, M.; Lancini, R.; and Tubaro, S. 2005.
\newblock Multiple description video coding for scalable and robust transmission over IP.
\newblock \emph{IEEE Transactions on circuits and systems for video technology}, 15(3): 321--334.

\bibitem[{He et~al.(2022)He, Yang, Peng, Ma, Qin, and Wang}]{he2022elic}
He, D.; Yang, Z.; Peng, W.; Ma, R.; Qin, H.; and Wang, Y. 2022.
\newblock Elic: Efficient learned image compression with unevenly grouped space-channel contextual adaptive coding.
\newblock In \emph{Proceedings of the IEEE/CVF Conference on Computer Vision and Pattern Recognition}, 5718--5727.

\bibitem[{Hu et~al.(2023)Hu, Ghosh, Liu, Zhang, and Shroff}]{hu2023corel}
Hu, X.; Ghosh, A.; Liu, X.; Zhang, Z.-L.; and Shroff, N. 2023.
\newblock COREL: Constrained Reinforcement Learning for Video Streaming ABR Algorithm Design Over mmWave 5G.
\newblock In \emph{2023 IEEE International Workshop Technical Committee on Communications Quality and Reliability (CQR)}, 1--6. IEEE.

\bibitem[{Hu et~al.(2021)Hu, Pan, Wang, Zhang, and Shirmohammadi}]{hu2021multiple}
Hu, X.; Pan, Y.; Wang, Y.; Zhang, L.; and Shirmohammadi, S. 2021.
\newblock Multiple description coding for best-effort delivery of light field video using GNN-based compression.
\newblock \emph{IEEE Transactions on Multimedia}, 25: 690--705.

\bibitem[{Hu et~al.(2022)Hu, Lu, Guo, Liu, Jiang, and Xu}]{hu2022coarse}
Hu, Z.; Lu, G.; Guo, J.; Liu, S.; Jiang, W.; and Xu, D. 2022.
\newblock Coarse-to-fine deep video coding with hyperprior-guided mode prediction.
\newblock In \emph{Proceedings of the IEEE/CVF Conference on Computer Vision and Pattern Recognition}, 5921--5930.

\bibitem[{Hu, Lu, and Xu(2021)}]{2021-FVC}
Hu, Z.; Lu, G.; and Xu, D. 2021.
\newblock FVC: A new framework towards deep video compression in feature space.
\newblock In \emph{Proceedings of the IEEE/CVF Conference on Computer Vision and Pattern Recognition}, 1502--1511.

\bibitem[{Kazemi, Shirmohammadi, and Sadeghi(2014)}]{kazemi2014review}
Kazemi, M.; Shirmohammadi, S.; and Sadeghi, K.~H. 2014.
\newblock A review of multiple description coding techniques for error-resilient video delivery.
\newblock \emph{Multimedia Systems}, 20: 283--309.

\bibitem[{Kwan et~al.(2024)Kwan, Gao, Zhang, Gower, and Bull}]{kwan2024hinerv}
Kwan, H.~M.; Gao, G.; Zhang, F.; Gower, A.; and Bull, D. 2024.
\newblock HiNeRV: Video Compression with Hierarchical Encoding-based Neural Representation.
\newblock \emph{Advances in Neural Information Processing Systems}, 36.

\bibitem[{Le et~al.(2023)Le, Antonini, Lambert, and Alioua}]{le2023multiple}
Le, T.~H.; Antonini, M.; Lambert, M.; and Alioua, K. 2023.
\newblock Multiple description video coding for real-time applications using HEVC.
\newblock In \emph{2023 IEEE International Conference on Image Processing (ICIP)}, 2580--2584. IEEE.

\bibitem[{Le, Pic, and Antonini(2023)}]{le2023inr}
Le, T.~H.; Pic, X.; and Antonini, M. 2023.
\newblock INR-MDSQC: Implicit Neural Representation Multiple Description Scalar Quantization for robust image Coding.
\newblock In \emph{2023 IEEE 25th International Workshop on Multimedia Signal Processing (MMSP)}, 1--6. IEEE.

\bibitem[{Li, Li, and Lu(2021{\natexlab{a}})}]{2021-Deep}
Li, J.; Li, B.; and Lu, Y. 2021{\natexlab{a}}.
\newblock Deep contextual video compression.
\newblock \emph{Advances in Neural Information Processing Systems}, 34: 18114--18125.

\bibitem[{Li, Li, and Lu(2021{\natexlab{b}})}]{li2021deep}
Li, J.; Li, B.; and Lu, Y. 2021{\natexlab{b}}.
\newblock Deep contextual video compression.
\newblock \emph{Advances in Neural Information Processing Systems}, 34: 18114--18125.

\bibitem[{Li, Li, and Lu(2023)}]{li2023neural}
Li, J.; Li, B.; and Lu, Y. 2023.
\newblock Neural video compression with diverse contexts.
\newblock In \emph{Proceedings of the IEEE/CVF conference on computer vision and pattern recognition}, 22616--22626.

\bibitem[{Li et~al.(2023{\natexlab{a}})Li, Zhang, Liu, Tan, Peng, and Lu}]{li2023ca++}
Li, Q.; Zhang, Z.; Liu, Y.; Tan, Z.; Peng, C.; and Lu, S. 2023{\natexlab{a}}.
\newblock CA++: Enhancing Carrier Aggregation Beyond 5G.
\newblock In \emph{Proceedings of the 29th Annual International Conference on Mobile Computing and Networking}, 1--14.

\bibitem[{Li et~al.(2023{\natexlab{b}})Li, Chang, Mishra, Zhang, Katabi, and Krishnan}]{li2023mage}
Li, T.; Chang, H.; Mishra, S.; Zhang, H.; Katabi, D.; and Krishnan, D. 2023{\natexlab{b}}.
\newblock Mage: Masked generative encoder to unify representation learning and image synthesis.
\newblock In \emph{Proceedings of the IEEE/CVF Conference on Computer Vision and Pattern Recognition}, 2142--2152.

\bibitem[{Lin et~al.(2022)Lin, Jia, Zhang, Wang, Ma, and Gao}]{2022-decompose}
Lin, K.; Jia, C.; Zhang, X.; Wang, S.; Ma, S.; and Gao, W. 2022.
\newblock DMVC: Decomposed motion modeling for learned video compression.
\newblock \emph{IEEE Transactions on Circuits and Systems for Video Technology}.

\bibitem[{Liu et~al.(2024)Liu, Wu, Wang, Liu, Yap, and Chau}]{liu2024bitstream}
Liu, T.; Wu, K.; Wang, Y.; Liu, W.; Yap, K.-H.; and Chau, L.-P. 2024.
\newblock Bitstream-Corrupted Video Recovery: A Novel Benchmark Dataset and Method.
\newblock \emph{Advances in Neural Information Processing Systems}, 36.

\bibitem[{Lu et~al.(2019)Lu, Ouyang, Xu, Zhang, Cai, and Gao}]{lu2019dvc}
Lu, G.; Ouyang, W.; Xu, D.; Zhang, X.; Cai, C.; and Gao, Z. 2019.
\newblock Dvc: An end-to-end deep video compression framework.
\newblock In \emph{Proceedings of the IEEE/CVF Conference on Computer Vision and Pattern Recognition}, 11006--11015.

\bibitem[{Mentzer, Agustson, and Tschannen(2023)}]{m2t}
Mentzer, F.; Agustson, E.; and Tschannen, M. 2023.
\newblock M2t: Masking transformers twice for faster decoding.
\newblock In \emph{Proceedings of the IEEE/CVF International Conference on Computer Vision}, 5340--5349.

\bibitem[{Mentzer et~al.(2022)Mentzer, Toderici, Minnen, Hwang, Caelles, Lucic, and Agustsson}]{2022-VCT}
Mentzer, F.; Toderici, G.; Minnen, D.; Hwang, S.-J.; Caelles, S.; Lucic, M.; and Agustsson, E. 2022.
\newblock Vct: A video compression transformer.
\newblock \emph{arXiv preprint arXiv:2206.07307}.

\bibitem[{Mercat, Viitanen, and Vanne(2020)}]{mercat2020uvg}
Mercat, A.; Viitanen, M.; and Vanne, J. 2020.
\newblock UVG dataset: 50/120fps 4K sequences for video codec analysis and development.
\newblock In \emph{Proceedings of the 11th ACM Multimedia Systems Conference}, 297--302.

\bibitem[{Minnen, Ball{\'e}, and Toderici(2018)}]{minnen2018joint}
Minnen, D.; Ball{\'e}, J.; and Toderici, G.~D. 2018.
\newblock Joint autoregressive and hierarchical priors for learned image compression.
\newblock \emph{Advances in neural information processing systems}, 31.

\bibitem[{Minnen and Singh(2020)}]{minnen2020channel}
Minnen, D.; and Singh, S. 2020.
\newblock Channel-wise autoregressive entropy models for learned image compression.
\newblock In \emph{2020 IEEE International Conference on Image Processing (ICIP)}, 3339--3343. IEEE.

\bibitem[{Narayanan et~al.(2020{\natexlab{a}})Narayanan, Ramadan, Carpenter, Liu, Liu, Qian, and Zhang}]{first5G}
Narayanan, A.; Ramadan, E.; Carpenter, J.; Liu, Q.; Liu, Y.; Qian, F.; and Zhang, Z.-L. 2020{\natexlab{a}}.
\newblock A First Look at Commercial 5G Performance on Smartphones.
\newblock In \emph{Proceedings of The Web Conference 2020}, WWW '20, 894–905. New York, NY, USA: Association for Computing Machinery.
\newblock ISBN 9781450370233.

\bibitem[{Narayanan et~al.(2020{\natexlab{b}})Narayanan, Ramadan, Mehta, Hu, Liu, Fezeu, Dayalan, Verma, Ji, Li et~al.}]{narayanan2020lumos5g}
Narayanan, A.; Ramadan, E.; Mehta, R.; Hu, X.; Liu, Q.; Fezeu, R.~A.; Dayalan, U.~K.; Verma, S.; Ji, P.; Li, T.; et~al. 2020{\natexlab{b}}.
\newblock Lumos5G: Mapping and predicting commercial mmWave 5G throughput.
\newblock In \emph{Proceedings of the ACM Internet Measurement Conference}, 176--193.

\bibitem[{Narayanan et~al.(2021)Narayanan, Zhang, Zhu, Hassan, Jin, Zhu, Zhang, Rybkin, Yang, Mao et~al.}]{narayanan2021variegated}
Narayanan, A.; Zhang, X.; Zhu, R.; Hassan, A.; Jin, S.; Zhu, X.; Zhang, X.; Rybkin, D.; Yang, Z.; Mao, Z.~M.; et~al. 2021.
\newblock A variegated look at 5G in the wild: performance, power, and QoE implications.
\newblock In \emph{Proceedings of the 2021 ACM SIGCOMM 2021 Conference}, 610--625.

\bibitem[{Radulovic et~al.(2009)Radulovic, Frossard, Wang, Hannuksela, and Hallapuro}]{radulovic2009multiple}
Radulovic, I.; Frossard, P.; Wang, Y.-K.; Hannuksela, M.~M.; and Hallapuro, A. 2009.
\newblock Multiple description video coding with H. 264/AVC redundant pictures.
\newblock \emph{IEEE Transactions on Circuits and Systems for Video Technology}, 20(1): 144--148.

\bibitem[{Ramadan et~al.(2021)Ramadan, Narayanan, Dayalan, Fezeu, Qian, and Zhang}]{5gaware}
Ramadan, E.; Narayanan, A.; Dayalan, U.~K.; Fezeu, R. A.~K.; Qian, F.; and Zhang, Z.-L. 2021.
\newblock Case for 5G-Aware Video Streaming Applications.
\newblock In \emph{Proceedings of the 1st Workshop on 5G Measurements, Modeling, and Use Cases}, 5G-MeMU '21, 27–34. New York, NY, USA: Association for Computing Machinery.
\newblock ISBN 9781450386364.

\bibitem[{Rippel et~al.(2021)Rippel, Anderson, Tatwawadi, Nair, Lytle, and Bourdev}]{rippel2021elf}
Rippel, O.; Anderson, A.~G.; Tatwawadi, K.; Nair, S.; Lytle, C.; and Bourdev, L. 2021.
\newblock Elf-vc: Efficient learned flexible-rate video coding.
\newblock In \emph{Proceedings of the IEEE/CVF International Conference on Computer Vision}, 14479--14488.

\bibitem[{Rochman et~al.(2023)Rochman, Ye, Zhang, and Ghosh}]{DynSPAN24-Muhamed-Wei}
Rochman, M.~I.; Ye, W.; Zhang, Z.-L.; and Ghosh, M. 2023.
\newblock A Comprehensive Real-World Evaluation of 5G Improvements over 4G in Low-and Mid-Bands.
\newblock \emph{arXiv preprint arXiv:2312.00957}.

\bibitem[{Shirani(2006)}]{shirani2006content}
Shirani, S. 2006.
\newblock Content-based multiple description image coding.
\newblock \emph{IEEE transactions on multimedia}, 8(2): 411--419.

\bibitem[{Stevens(1997)}]{stevens1997rfc2001}
Stevens, W. 1997.
\newblock RFC2001: TCP slow start, congestion avoidance, fast retransmit, and fast recovery algorithms.

\bibitem[{Theis et~al.(2022)Theis, Shi, Cunningham, and Husz{\'a}r}]{theis2022lossy}
Theis, L.; Shi, W.; Cunningham, A.; and Husz{\'a}r, F. 2022.
\newblock Lossy image compression with compressive autoencoders.
\newblock In \emph{International conference on learning representations}.

\bibitem[{Tillo and Olmo(2004)}]{tillo2004low}
Tillo, T.; and Olmo, G. 2004.
\newblock Low complexity pre postprocessing multiple description coding for video streaming.
\newblock In \emph{Proceedings. 2004 International Conference on Information and Communication Technologies: From Theory to Applications, 2004.}, 519--520. IEEE.

\bibitem[{Wang et~al.(2016)Wang, Gan, Hu, Lin, Jin, Song, Wang, Katsavounidis, Aaron, and Kuo}]{wang2016mcl}
Wang, H.; Gan, W.; Hu, S.; Lin, J.~Y.; Jin, L.; Song, L.; Wang, P.; Katsavounidis, I.; Aaron, A.; and Kuo, C.-C.~J. 2016.
\newblock MCL-JCV: a JND-based H. 264/AVC video quality assessment dataset.
\newblock In \emph{2016 IEEE international conference on image processing (ICIP)}, 1509--1513. IEEE.

\bibitem[{Wang et~al.(2001)Wang, Orchard, Vaishampayan, and Reibman}]{wang2001multiple}
Wang, Y.; Orchard, M.~T.; Vaishampayan, V.; and Reibman, A.~R. 2001.
\newblock Multiple description coding using pairwise correlating transforms.
\newblock \emph{IEEE Transactions on Image Processing}, 10(3): 351--366.

\bibitem[{Wang, Simoncelli, and Bovik(2003)}]{wang2003multiscale}
Wang, Z.; Simoncelli, E.~P.; and Bovik, A.~C. 2003.
\newblock Multiscale structural similarity for image quality assessment.
\newblock In \emph{The Thrity-Seventh Asilomar Conference on Signals, Systems \& Computers, 2003}, volume~2, 1398--1402. Ieee.

\bibitem[{Wicker and Bhargava(1999)}]{wicker1999reed}
Wicker, S.~B.; and Bhargava, V.~K. 1999.
\newblock \emph{Reed-Solomon codes and their applications}.
\newblock John Wiley \& Sons.

\bibitem[{Xiang, Tian, and Zhang(2022)}]{xiang2022mimt}
Xiang, J.; Tian, K.; and Zhang, J. 2022.
\newblock Mimt: Masked image modeling transformer for video compression.
\newblock In \emph{The Eleventh International Conference on Learning Representations}.

\bibitem[{Xue et~al.(2019)Xue, Chen, Wu, Wei, and Freeman}]{xue2019video}
Xue, T.; Chen, B.; Wu, J.; Wei, D.; and Freeman, W.~T. 2019.
\newblock Video enhancement with task-oriented flow.
\newblock \emph{International Journal of Computer Vision}, 127: 1106--1125.

\bibitem[{Yap{\i}c{\i} et~al.(2008)Yap{\i}c{\i}, Demir, Ert{\"u}rk, and Urhan}]{yapici2008downsampling}
Yap{\i}c{\i}, Y.; Demir, B.; Ert{\"u}rk, S.; and Urhan, O. 2008.
\newblock Downsampling-based multiple description image coding using optimal filtering.
\newblock \emph{Journal of Electronic Imaging}, 17(3): 033018--033018.

\bibitem[{Ye et~al.(2024)Ye, Hu, Sleder, Zhang, Dayalan, Hassan, Fezeu, Jajoo, Lee, Ramadan et~al.}]{ye2024dissecting}
Ye, W.; Hu, X.; Sleder, S.; Zhang, A.; Dayalan, U.~K.; Hassan, A.; Fezeu, R.~A.; Jajoo, A.; Lee, M.; Ramadan, E.; et~al. 2024.
\newblock Dissecting Carrier Aggregation in 5G Networks: Measurement, QoE Implications and Prediction.
\newblock In \emph{Proceedings of the ACM SIGCOMM 2024 Conference}, 340--357.

\bibitem[{Yu et~al.(2023)Yu, Cheng, Sohn, Lezama, Zhang, Chang, Hauptmann, Yang, Hao, Essa et~al.}]{yu2023magvit}
Yu, L.; Cheng, Y.; Sohn, K.; Lezama, J.; Zhang, H.; Chang, H.; Hauptmann, A.~G.; Yang, M.-H.; Hao, Y.; Essa, I.; et~al. 2023.
\newblock Magvit: Masked generative video transformer.
\newblock In \emph{Proceedings of the IEEE/CVF Conference on Computer Vision and Pattern Recognition}, 10459--10469.

\bibitem[{Zhao et~al.(2018)Zhao, Bai, Wang, and Zhao}]{zhao2018multiple}
Zhao, L.; Bai, H.; Wang, A.; and Zhao, Y. 2018.
\newblock Multiple description convolutional neural networks for image compression.
\newblock \emph{IEEE Transactions on Circuits and Systems for Video Technology}, 29(8): 2494--2508.

\bibitem[{Zhao et~al.(2022)Zhao, Zhang, Bai, Wang, and Zhao}]{zhao2022lmdc}
Zhao, L.; Zhang, J.; Bai, H.; Wang, A.; and Zhao, Y. 2022.
\newblock LMDC: Learning a multiple description codec for deep learning-based image compression.
\newblock \emph{Multimedia Tools and Applications}, 81(10): 13889--13910.

\bibitem[{Zheng et~al.(2021)Zheng, Ma, Liu, Yang, Li, Zhang, Zhang, Shi, Chen, Li et~al.}]{zheng2021xlink}
Zheng, Z.; Ma, Y.; Liu, Y.; Yang, F.; Li, Z.; Zhang, Y.; Zhang, J.; Shi, W.; Chen, W.; Li, D.; et~al. 2021.
\newblock Xlink: Qoe-driven multi-path quic transport in large-scale video services.
\newblock In \emph{Proceedings of the 2021 ACM SIGCOMM 2021 Conference}, 418--432.

\end{thebibliography}
\clearpage
\appendix
\newpage
\section{Appendix / supplemental material}

\subsection{Real-world Network Traces}\label{a:5g_trace}
\begin{figure}[!ht]
    \centering
    \includegraphics[width=.48\linewidth]{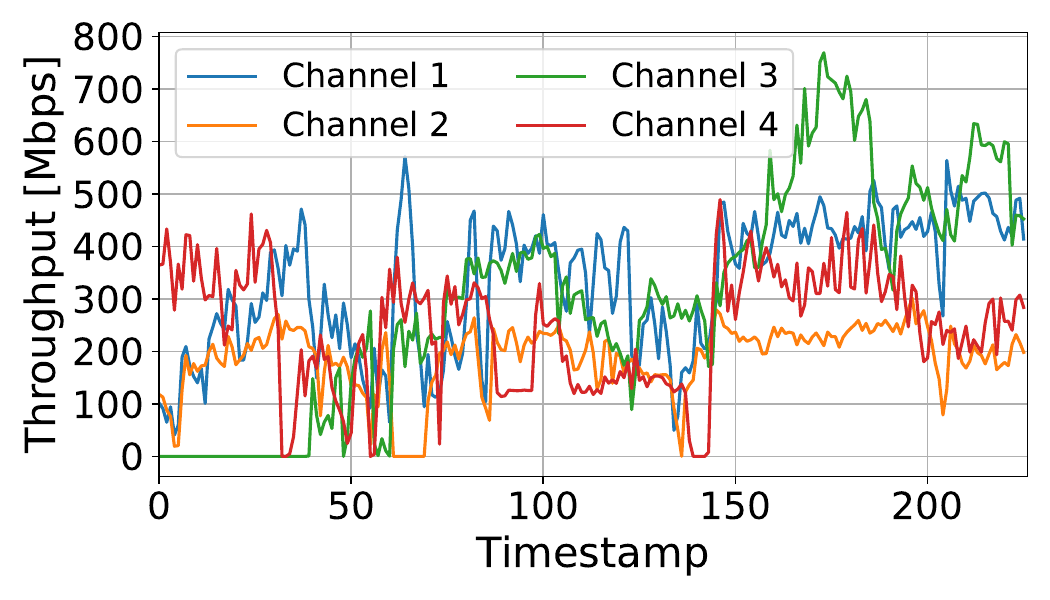}
    \includegraphics[width=.48\linewidth]{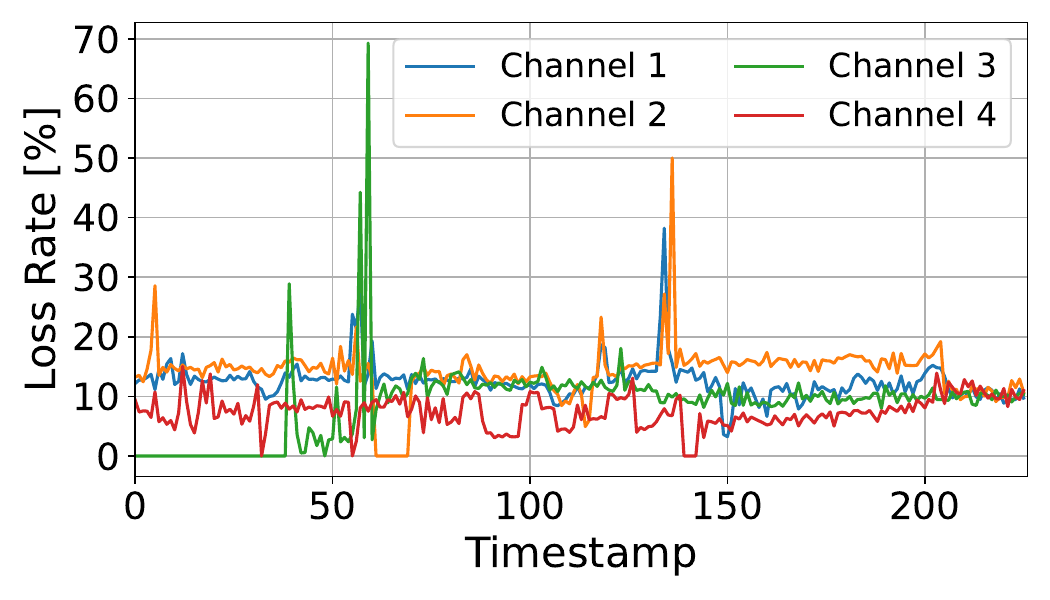}
    \vspace{-4mm}
    \caption{A representative sample of network traces showcasing dynamic throughput and transmission loss rates over time.}
    \label{f:dynamic_traces}
\end{figure}
Fig.~\ref{f:dynamic_traces} presents a sample of real-world network traces~\cite{ye2024dissecting}. The throughput and corresponding packet drop rates are highly dynamic over time. No channel consistently dominates and packet drop rate bursts for each channel occur at different times.

\subsection{Implementation Details} \label{a:implementation}
We implement \algorithmname on top of M2T~\cite{m2t} and VCT ~\cite{2022-VCT}, two recent works utilizing masked and unmasked transformers for image and video compression. 
To achieve various bitrate control, we optimize the training loss for six values of $\lambda$, ranging from 0.0001 to 1. We use the linearly decaying learning rate schedule with warmup. The base learning rate is $10^{-4}$. We warmup for 10k steps, keep the learning rate constant until reaching $90\%$ of the training process. Then we linearly decay the learning rate to $10^{-5}$. We use 12 QLDS masking schedules, the parameter $\alpha$ of which is 2.2, for iterative entropy encoding and decoding. 
All experiments are conducted on Nvidia A6000 GPUs and independently run three times.

\subsection{Inference example with loss}

\begin{figure*}[!ht]
    \centering
    \subfigure[\shortstack{ \protect\\Original }]{\includegraphics[width=.24\linewidth]{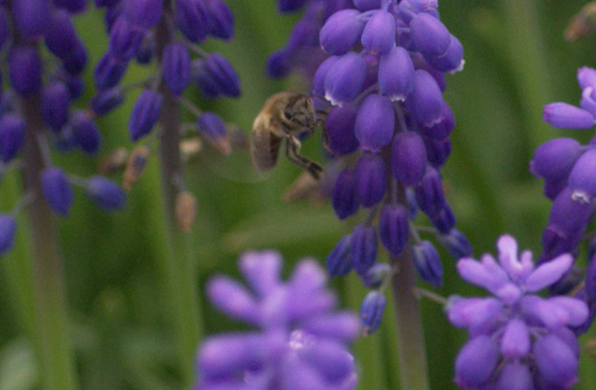}}
    \subfigure[\shortstack{ PSNR: 32.11  MS-SSIM: 0.95\protect\\DCVC-DC(w/o RTX)}]{\includegraphics[width=.24\linewidth]{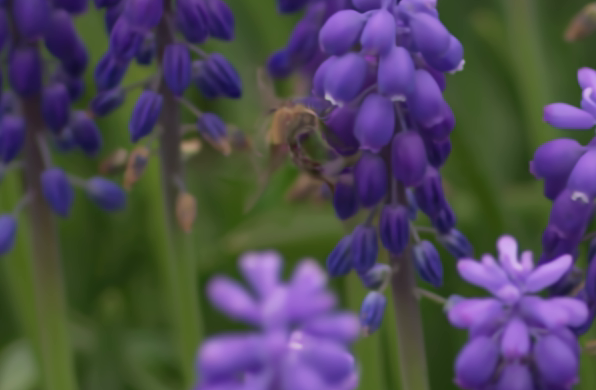}}
    \subfigure[\shortstack{ PSNR: 32.81  MS-SSIM: 0.95\protect\\DCVC-DC(RTX, RTT=10))}]{\includegraphics[width=.24\linewidth]{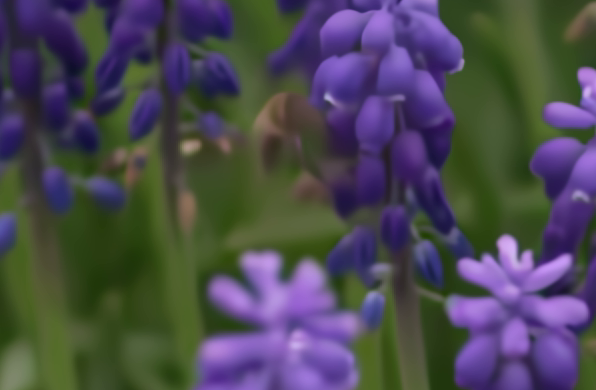}}
    \subfigure[\shortstack{PSNR: 33.68  MS-SSIM: 0.96\protect\\NeuralMDC(w/o RTX)}]{\includegraphics[width=.24\linewidth]{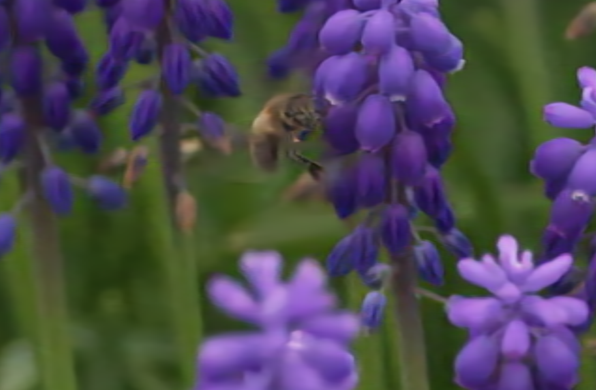}}\\
    \subfigure[\shortstack{\protect\\Original }]{\includegraphics[width=.24\linewidth]{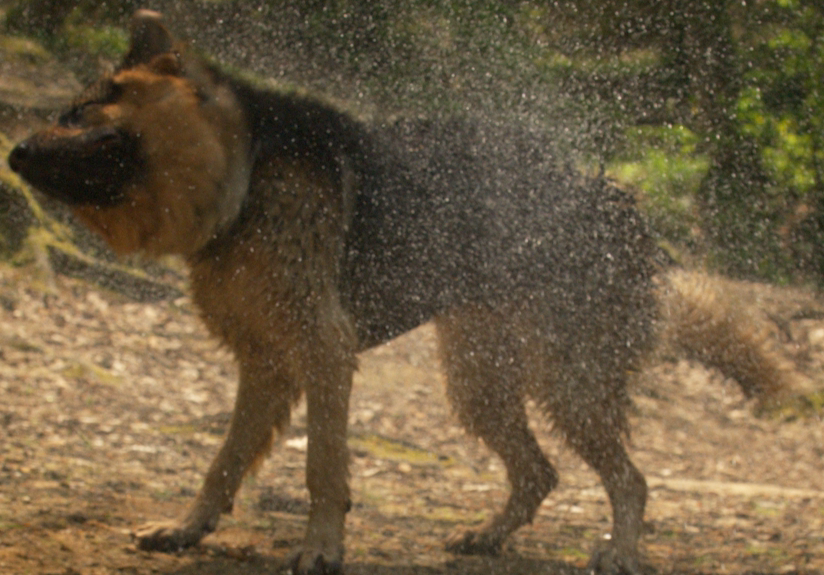}}
    \subfigure[\shortstack{ PSNR: 22.20  MS-SSIM: 0.70\protect\\DCVC-DC(w/o RTX)}]{\includegraphics[width=.24\linewidth]{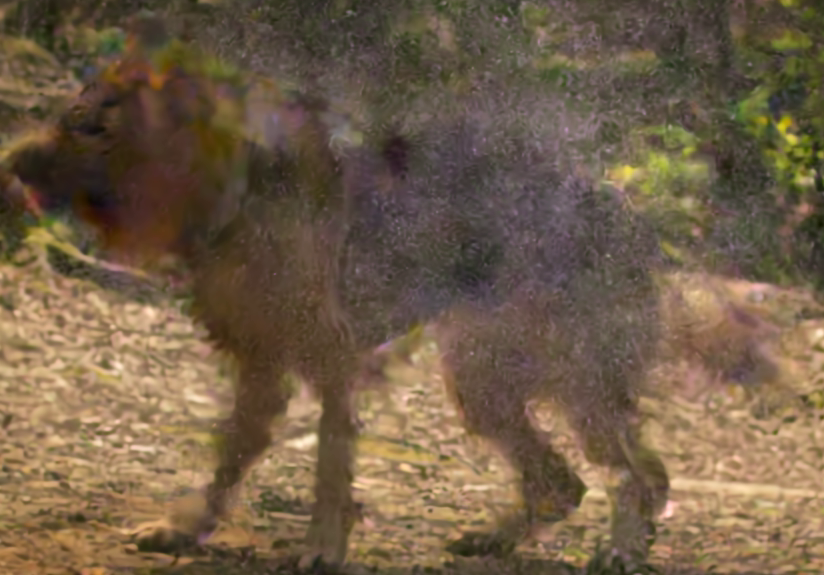}}
    \subfigure[\shortstack{ PSNR: 25.12  MS-SSIM: 0.67\protect\\DCVC-DC(RTX, RTT=10)}]{\includegraphics[width=.24\linewidth]{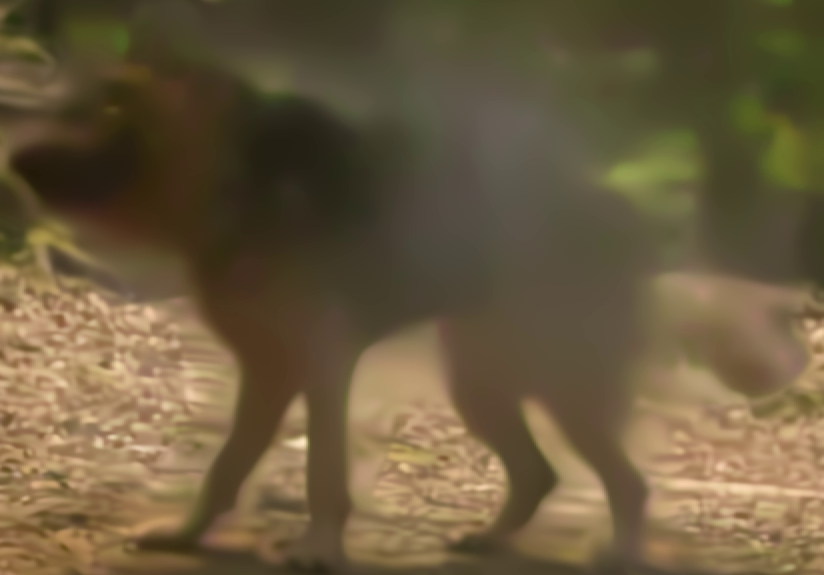}}
    \subfigure[\shortstack{PSNR: 29.92 MS-SSM: 0.88\\NeuralMDC(w/o RTX)}]{\includegraphics[width=.24\linewidth]{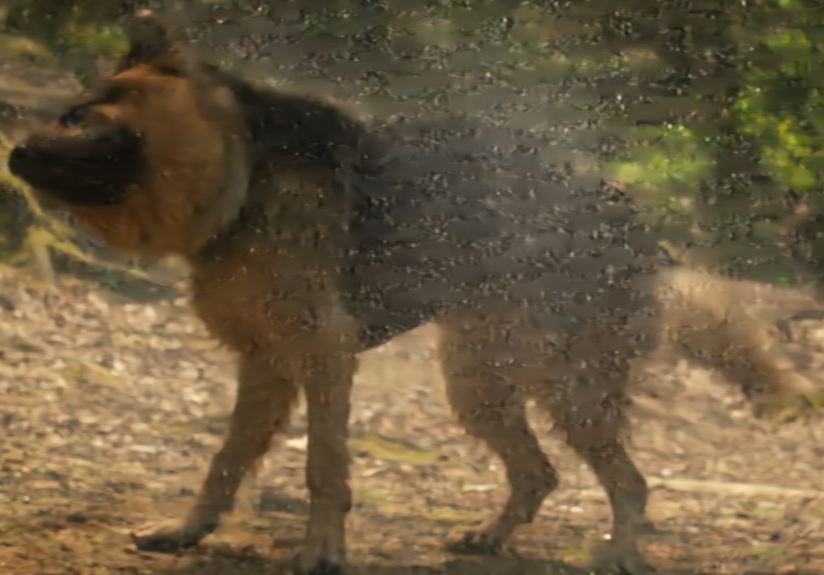}}\\
    \subfigure[\shortstack{ \protect\\Original }]{\includegraphics[width=.24\linewidth]{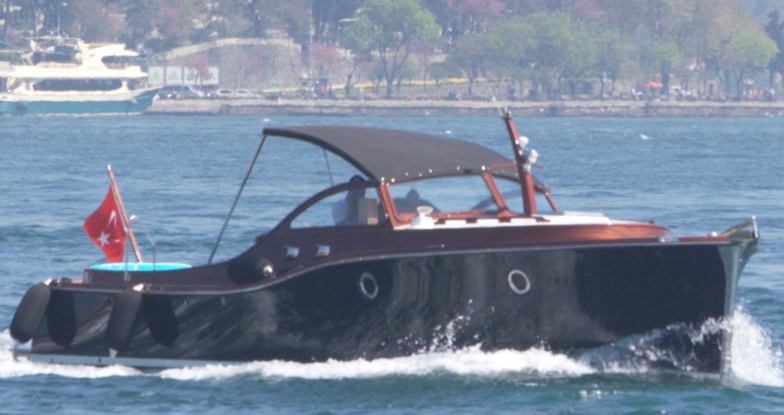}}
    \subfigure[\shortstack{ PSNR: 22.21  MS-SSIM: 0.74\protect\\DCVC-DC(w/o RTX)}]{\includegraphics[width=.24\linewidth]{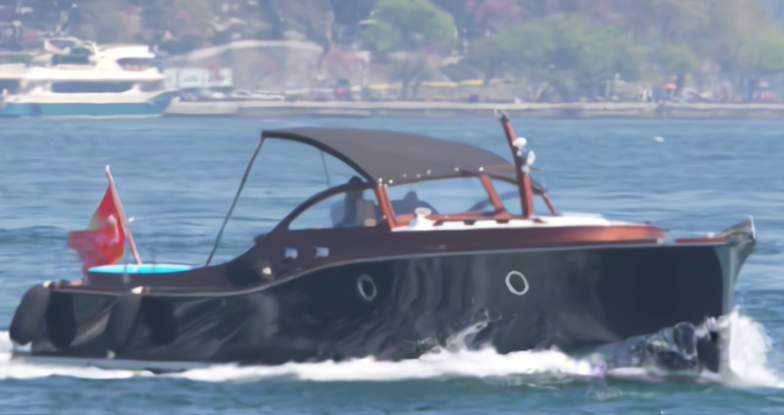}}
    \subfigure[\shortstack{ PSNR: 27.71  MS-SSIM: 0.88\protect\\DCVC-DC(RTX, RTT=10)}]{\includegraphics[width=.24\linewidth]{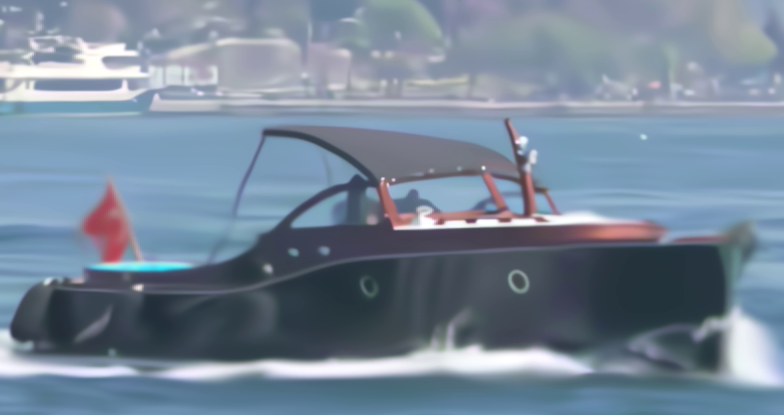}}
    \subfigure[\shortstack{PSNR: 32.85 MS-SSIM: 0.96\\NeuralMDC(w/o RTX)}]{\includegraphics[width=.24\linewidth]{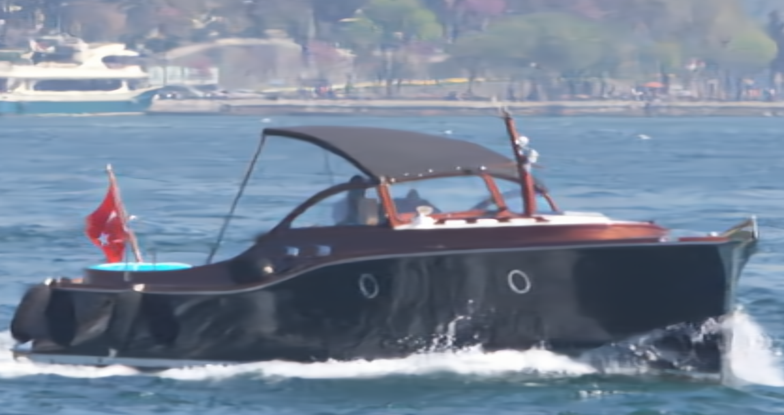}}\\
    \caption{Inference examples with 50\% packet loss of NeuralMDC and DCVC-DC under the same network bandwidth and transmission time: bpp(DCVC-DC w/o RTX)=0.0197, bpp(DCVC-DC RTX RTT10)=0.00663, bpp(NeuralMDC w/o RTX)=0.0177. (RTX means retransmission and RTT means network round-trip time )} 
    \label{f:reconstruct}
\end{figure*}

We present some reconstruction examples of \algorithmname and DCVC-DC under a representative 50\% packet loss in Figure \ref{f:reconstruct} together with the original frame. Also, the metric PSNR and MS-SSIM are attached at the bottom of each example. Clearly, the examples show the capacity of \algorithmname's superior reconstruction and loss resilience to a certain amount of loss.

\end{document}